\documentclass{article} 
\usepackage{iclr2025_conference,times}

\usepackage{amsmath,amsfonts,bm}

\def\eqref#1{equation~\ref{#1}}

\def\1{\bm{1}}

\DeclareMathAlphabet{\mathsfit}{\encodingdefault}{\sfdefault}{m}{sl}
\SetMathAlphabet{\mathsfit}{bold}{\encodingdefault}{\sfdefault}{bx}{n}

\usepackage{hyperref}
\usepackage{url}

\usepackage{graphicx}
\usepackage{booktabs}
\usepackage{multirow}
\usepackage{subfig}
\usepackage{color, xcolor}

\usepackage{amsmath}
\usepackage{amssymb}
\usepackage{mathtools}
\usepackage{amsthm}
\usepackage{pifont}
\usepackage{tcolorbox} 
\tcbuselibrary{breakable} 
\usepackage{soul} 
\usepackage{tikz} 
\usepackage{colortbl} 
\usepackage{adjustbox} 
\usepackage{wrapfig} 
\usepackage{enumitem} 
\usepackage{subcaption} 

\newcommand{\think}[1]{\textcolor{blue}{\texttt{\textbf{<think>}}} #1 \textcolor{blue}{\texttt{\textbf{</think>}}}}
\newcommand{\thought}[1]{\textcolor{cyan}{\texttt{\textbf{<begin\_of\_thought>}}} #1 \textcolor{cyan}{\texttt{\textbf{<end\_of\_thought>}}}}
\newcommand{\truesolution}[1]{\textcolor{green}{\texttt{\textbf{<begin\_of\_solution>}}} #1 \textcolor{green}{\texttt{\textbf{<end\_of\_solution>}}}}
\newcommand{\falsesolution}[1]{\textcolor{red}{\texttt{\textbf{<begin\_of\_solution>}}} #1 \textcolor{red}{\texttt{\textbf{<end\_of\_solution>}}}}

\definecolor{darkgreen}{rgb}{0.0, 0.5, 0.0}

\newcommand{\OurMODEL}{\textsc{LongPAS}}
\newcommand{\OurDataset}{\textsc{DeepReasonQA}}

\title{Incentivizing In-depth Reasoning over Long Contexts with Process Advantage Shaping}

\author{Miao Peng$^{1,2}$\thanks{$\;$Research done during internship at Tongyi Lab, Alibaba Group} \quad
Weizhou Shen$^{2}$ \quad
Nuo Chen$^{1}$ \quad
Chenliang Li$^{2}$ \quad
Ming Yan$^{2}$\thanks{$\;$Corresponding authors.} \quad
Jia Li$^{1}$\footnotemark[\value{footnote}] \\
$^{1}$The Hong Kong University of Science and Technology (Guangzhou)\\
$^{2}$Tongyi Lab, Alibaba Group \\
    \texttt{\{mpeng885, jialee\}@connect.hkust-gz.edu.cn} \\
    \texttt{ym119608@alibaba-inc.com}
}
\iclrfinalcopy 
\begin{document}

\maketitle

\begin{abstract}
Reinforcement Learning with Verifiable Rewards (RLVR) has proven effective in enhancing LLMs short-context reasoning, but its performance degrades in long‑context scenarios that require both precise grounding and robust long‑range reasoning. We identify the ``almost-there" phenomenon—trajectories that are largely correct but fail at the final step—in long-context reasoning RL and attribute this failure to two factors: (1) the lack of high reasoning density in long-context QA data that push LLMs beyond mere grounding toward sophisticated multi-hop reasoning; and (2) the loss of valuable learning signals during long-context RL training due to the indiscriminate penalization of partially correct trajectories with incorrect outcomes. To overcome this bottleneck, we propose \textbf{\OurDataset{}}, a KG-driven synthesis framework that controllably constructing high-difficulty, multi-hop long-context QA pairs with inherent reasoning chains. Building on this, we introduce Long-context Process Advantage Shaping (\textbf{\OurMODEL{}}), a simple yet effective method that performs fine-grained credit assignment by evaluating reasoning steps along \textit{Validity} and \textit{Relevance} dimensions, which captures critical learning signals from ``almost-there" trajectories. Experiments on three long-context reasoning benchmarks show that our approach substantially outperforms RLVR baselines and matches frontier LLMs while using far fewer parameters. Further analysis confirms the effectiveness of our methods in strengthening long-context reasoning while maintaining stable RL training.
\end{abstract}

\section{Introduction}

Reasoning over long contexts is a critical capability for modern large language models (LLMs), as many real‑world tasks—such as document understanding~\citep{LongBench, LongBench_v2} or agentic deep research~\citep{search_r1, tongyi_deepresearch}—require grounding information and perform complex reasoning across millions of tokens~\cite{RULER, LongReason, Frames}. While advanced LLMs have successfully employ RLVR (Reinforcement Learning with Verifiable Rewards) to enhance short-context complex reasoning skills~\cite{Deepseek_r1, does_RL}, performance still degrades significantly when confronted with long-context scenarios. Recent work has explored improving long‑context capability in RL by progressively scaling the context window~\citep{QwenLong_L1}, or by using data‑driven RL approaches to learn plan–retrieve–reason patterns~\citep{LoongRL}. However, being primarily focused on improving information grounding, these methods still struggle with in-depth reasoning over long-context documents.

A key reason is that they suffer from the limitation of current outcome‑based RLVR algorithms~\cite{GRPO, DAPO, reason_entropy}: These methods uniformly rely on sparse, outcome‑level rewards, making it inherently difficult to distinguish the common ``almost‑there" cases—instances where most reasoning steps are correct but the final answer is wrong. Such reward sparsity causes RLVR to ignore valuable learning signals contained in these partially correct trajectories, a problem that becomes especially severe in complex long‑context tasks where these cases are frequent. To empirically validate this, we examined Entity Coverage and Triple Coverage across varying uncertainty levels in Figure~\ref{fig:Analysis_on_Trajectory_Coverage}.
It reveals that in high-uncertainty samples (accuracy around 50\%), the model successfully anchors most critical entities but falters during information integration and final reasoning (Further analysis in Section~\ref{sec:analysis_steps_in_negative}). Despite their potential to provide valuable learning signals~\citep{RL_difficulty}, these ``almost-there" trajectories are effectively penalized as failures under sparse reward regimes.

While integrating process-level supervision appears to be a direct remedy for this sparsity, the unique characteristics of long-context tasks impose significant challenges: (1) \textbf{The Sparsity of Reasoning Density}: Unlike logic-heavy short-context tasks, key information in long documents is highly scattered. This ``reasoning-grounding imbalance" makes it difficult for LLMs to incentivize high-quality reasoning patterns without being overwhelmed by contextual noise. Besides, the vast search space of long-form text hinders effective construction of high-quality, step-level data with explicit dependencies. (2) \textbf{The Indeterminacy of Credit Assignment}: Many steps involve mere context grounding or document browsing, making it impractical to directly pinpoint which specific step or reasoning hop contributed to an "almost-there" failure, resulting in difficult and noisy fine-grained supervision signal assignment.
These limitations collectively raise the following key questions:

\begin{wrapfigure}[22]{r}{0.42\textwidth}
    \centering
    \vspace{-0.6cm}
    {\includegraphics[width=0.40\textwidth]{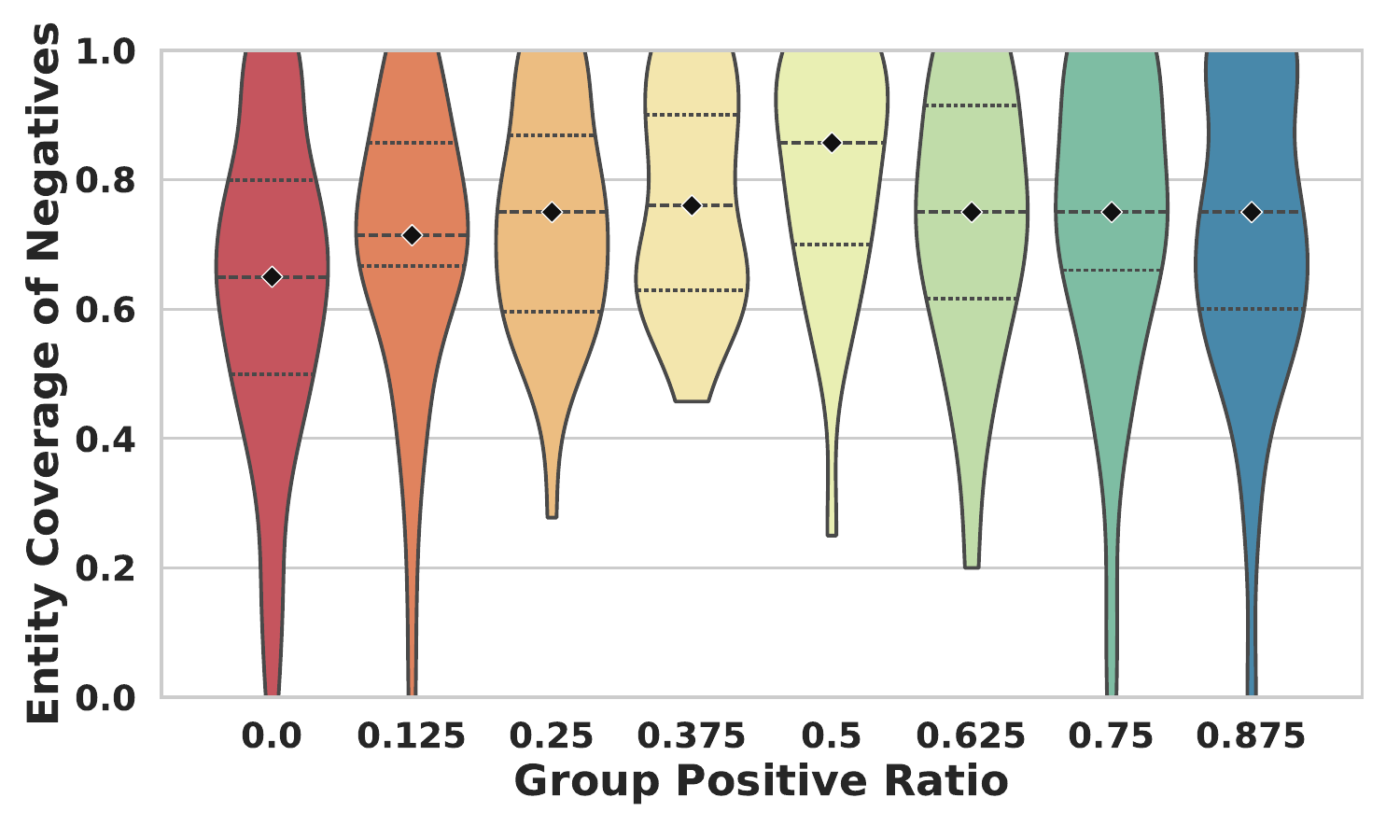} \label{Entity_Coverage}}
    {\includegraphics[width=0.40\textwidth]{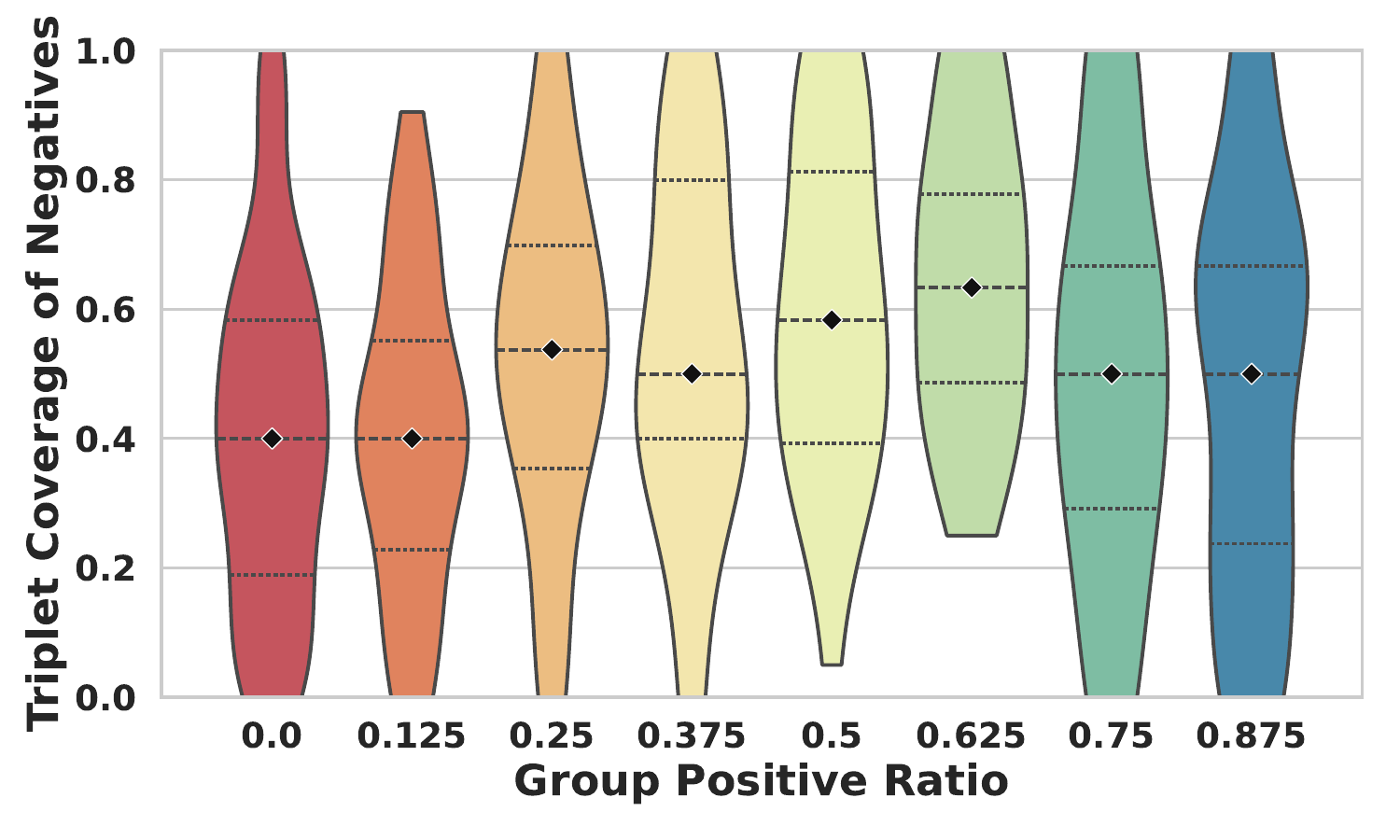} \label{Teiplet_Coverage}}
    
    \caption{Entity \& Triplet Coverage between negative rollouts and GT reasoning chains on FRAMES~\citep{Frames}.}
    \label{fig:Analysis_on_Trajectory_Coverage}
\end{wrapfigure}

\begin{enumerate}
    \item How to construct long-context QA data that possesses both ``high reasoning density" and ``explicit step dependencies," along with reliable step-level supervision signals?
    \item How can these step-level supervision signals be utilized to achieve fine-grained credit assignment and mitigate reward sparsity within the RL training process?
\end{enumerate}

To resolve these problems, we propose to shift the paradigm from mining supervision in noisy natural data to constructing supervision via controlled synthesis. We first propose KG-driven long-range multi-hop QA synthesis framework: We automatically extract and construct high-difficulty, complex multi-hop QA dataset \textbf{\OurDataset{}} from sparse long documents at scale, covering critical multi-hop reasoning scenarios. Besides, the controllable construction process inherently produces explicit reasoning chains alongside QA pair, which offers fine-grained step-level supervision. To enable effective RL training, we introduce Long-context Process Advantage Shaping (\textbf{\OurMODEL{}}) to improve credit assignment in long-context RL. \OurMODEL{} leverages reference reasoning chains to provide step-level supervision signals along \textit{Validity} and \textit{Relevance} dimensions, which are then utilized to assign a distinct advantage reweighting coefficient upon each token in a trajectory. This fine‑grained advantage shaping enables more precise credit assignment across the entire reasoning trajectory, preventing the collapse of ``almost-there" samples.

Our comprehensive evaluation on both in-domain and out-of-domain long-context multi-hop reasoning benchmarks demonstrates the high quality of \OurDataset{} and the effectiveness of \OurMODEL{}, surpassing strong RLVR baselines across multiple LLM series and achieving performance comparable to frontier LLMs with substantially larger parameters. Critically, beyond achieving superior accuracy, \OurMODEL{} also induces more effective reasoning policies that exhibit more precise and accurate information grounding, as well as more complex reasoning over long-context documents. Further analyses validate that our core algorithms support stable RL training and improve step-wise reasoning validity in challenging long-context reasoning scenarios.

\section{Preliminary}

\paragraph{Long-context RLVR}
Given a question $Q$ and a long context $C$ including a set of documents $c_1, c_2, ..., c_n$, the standard long-context RLVR (Reinforcement Learning with Verifiable Reward) framework aims to optimize a policy $\pi_{\theta}(y \mid C, Q)$ that generates the answer $A$ to the question. The objective is to maximize the expected verifiable reward $r_{\text{ans}}(y)$, which typically evaluates the correctness of the final answer through verifiable rules or preference reward model:
$$
J_{\text{ans}}(\theta) = \mathbb{E}_{(C, Q) \sim \mathcal{D}} \left[ \mathbb{E}_{y \sim \pi_{\theta}(y \mid C, Q)}[r_{\text{ans}}(y)] \right],
$$
where $D$ denotes the training set.

\paragraph{Group Relative Policy Optimization}
Group Relative Policy Optimization (GRPO)~\citep{GRPO} is a variant of Proximal Policy Optimization (PPO)~\citep{PPO}. GRPO maintains the current policy $\pi_{\theta}$ and a reference ``old" policy $\pi_{old}$. Given a prompt $(C, Q) \sim \mathcal{D}$, GRPO first samples a group of $N$ candidate answers $\boldsymbol{y} = \{y_1, \dots, y_G\}$ from the old policy $\pi_{old}(y \mid C, Q)$. For each answer $y_i \in \boldsymbol{y}$, the verifiable reward $r_i = r_{\text{ans}}(y_i)$ is computed. GRPO estimates the advantage using the relative quality of answers within this group.
The normalized advantage $\hat{A}_i$ for the $i$-th answer is then defined as:
\begin{equation*}
\hat{A}_i = \frac{r_i - \text{mean}(\{r_1,r_2,...,r_n\})}{\text{std}(\{r_1,r_2,...,r_n\})}.
\end{equation*}
The policy $\pi_{\theta}$ is optimized by maximizing the following objective function:
\begin{equation*}
J_{\text{GRPO}}(\theta) = \hat{\mathbb{E}}_{(C, Q), \boldsymbol{y}} \left[ \frac{1}{N} \sum_{i=1}^{N} f_{\epsilon} \left( \rho_i(\theta), \hat{A}_i \right) \right] - \beta \cdot \hat{\mathbb{E}}_{(C, Q)} \left[ \mathbb{D}_{KL}[\pi_{\theta}(\cdot \mid C, Q) \| \pi_{old}(\cdot \mid C, Q)] \right],
\end{equation*}
where $\hat{\mathbb{E}}$ denotes the empirical expectation over samples from $\mathcal{D}$ and $\boldsymbol{y} \sim \pi_{old}^N$. The term $\rho_i(\theta) = \frac{\pi_{\theta}(y_i \mid C, Q)}{\pi_{old}(y_i \mid C, Q)}$ represents the probability ratio. $f_{\epsilon}(x, A) = \min(xA, \text{clip}(x, 1-\epsilon, 1+\epsilon)A)$ is the standard PPO clipping function, and $\beta$ is a hyper-parameter controlling the KL divergence penalty.

\section{Methodology}

\subsection{The Challenge of Almost-there Phenomenon in Long-Context Reasoning}\label{sec:analysis_steps_in_negative}
Sparse reward signals often hinder RLVR when tackling complex long-context tasks. Given that these long trajectories consist of essential reasoning components like retrieval and information synthesis, they frequently harbor correct sub-steps even when the final output is wrong. Therefore, indiscriminate penalization based solely on final failures can be counterproductive, as it risks suppressing valid reasoning behaviors and undermining overall performance.

To quantify the prevalence of accurate reasoning steps within the trajectories of long-context multi-hop problems, we conduct an empirical analysis on FRAMES dataset~\citep{Frames} using Qwen3-4B model. Specifically, we leverage Gemini-2.5-Pro to identify the key ground-truth reasoning chains $P$ for each QA pair $(Q,A)$ in FRAMES. They are formally structured as chained triplets $(s_i,r_i,o_i)$, such as (DeMar DeRozan)$-$[played for]$\to$(San Antonio Spurs)$-$[competed in the]$\to$(Western Conference)... For each question, we perform N rollouts ${T_1, T_2,...,T_N}$ and categorize them based on the ratio of positive outcomes. Subsequently, we calculate the entity coverage and triplet coverage between each negative rollout (trajectories with false outcome), denoted as $T_f$, and its corresponding ground-truth reasoning chain $P$ using the following formulas:

\textbf{Entity Coverage} measures the proportion of correct entities in $P$ that are present in $T_f$, where $\mathcal{E}(T_f)$ and $\mathcal{E}(P)$ are the sets of entities in $T_f$ and $P$, respectively:
\begin{equation*}
\text{Entity Coverage}(T_f, P) = \frac{|\mathcal{E}(T_f) \cap \mathcal{E}(P)|}{|\mathcal{E}(P)|},
\end{equation*}

\textbf{Triplet Coverage} assessment relies on an LLM (LLM-as-a-judge) to evaluate the correctness of each step in $T_f$. We define $\mathcal{R}(T_f)$ as the set of triplets in $T_f$, and $\mathbb{I}_{Judge}(R, P)$ as an indicator function that equals 1 if the LLM judge deems the triplet $R \in \mathcal{R}(T_f)$ to be a valid and correct reasoning step present in $P$, and 0 otherwise. The Triplet Coverage is then calculated as:
\begin{equation*}
\text{Triplet Coverage}(T_f, P) = \frac{\sum_{R \in \mathcal{R}(T_f)} \mathbb{I}_{Judge}(R, P)}{|\mathcal{R}(P)|}.
\end{equation*}

We report the distributions of Entity \& Triplet Coverage between negative rollouts and ground-truth reasoning chains in Figure~\ref{fig:Analysis_on_Trajectory_Coverage}, and the empirical results reveal that the model's ``errors" are not uniform. As the Group Positive Ratio increases, suggesting decreased question difficulty, both Entity and Triplet Coverage of negative rollouts consistently improve, indicating that even failed trajectories are often containing high-ratio true reasoning steps.
Furthermore, the significantly higher Entity Coverage compared to Triplet Coverage demonstrates that while the model has a strong ability of information grounding, its capability to organize a strict logical chain remains a greater challenge.
Crucially, we observe that the coverage scores peak and remain above the overall average when the Group Positive Ratio is around 50\%. These samples represent questions where the model is on the boundary of success and failure, making them particularly valuable for RL training~\citep{RL_difficulty}. This finding underscores the necessity of implementing reward mechanisms that avoid incorrectly penalizing the correct sub-steps present within these ``almost-there" trajectories.

\begin{wrapfigure}[24]{r}{0.42\textwidth}
    \centering
    \vspace{-0.5cm}
    {\includegraphics[width=0.40\textwidth]{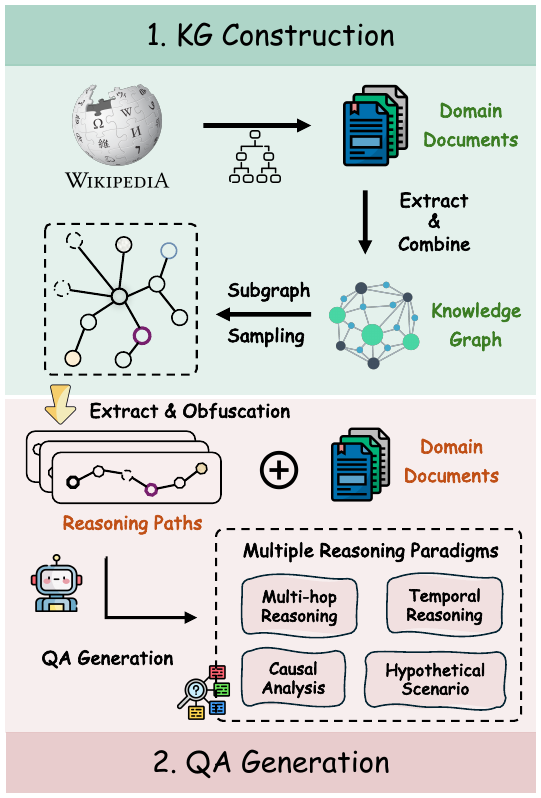}}
    
    \caption{Overall pipeline of the knowledge-guided long-context multi-hop QA synthesis framework.}
    \label{fig:QA_Synthesis_Pipeline}
\end{wrapfigure}

\subsection{Multi-hop Reasoning QA Synthesis}\label{sec:QA_synthesis}
To address the scarcity of high-reasoning-density QA pairs that necessitate long-range reasoning over long-context documents, we propose a Knowledge-Guided Long-Context Multi-hop QA Synthesis Framework. As shown in Figure~\ref{fig:QA_Synthesis_Pipeline}. This framework automates the extraction and construction of highly complex, multi-hop QA pairs from sparse, noisy, and long documents, thereby bypassing the need for manual data labeling. Crucially, it generates high-quality reasoning chains to accompany these complex questions, offering explicit dependency paths for stepwise supervision signals.

\paragraph{KG Construction} The data collection process involves aggregating Wikipedia page documents according to the hierarchical catalog, spanning diverse knowledge domains and subcategories including Sports, History, Science, Health, Biology, etc. We then perform triplet extraction from single documents and consolidate these into an initial Knowledge Graph $G$~\citep{KGGen}. A more complex, cross-document knowledge graph $G_d$ is subsequently formed via domain-level aggregation. To enhance the quality and cohesion of the graph, the cross-document knowledge graph $G_d$ is further refined by methods including entity and relation clustering.

\paragraph{Reasoning Path Sampling} To generate challenging multi-hop paths, we first sample relation-relevant subgraphs centered on target entities within the domain-specific knowledge graph. Long-range reasoning paths are subsequently derived using sampling strategies such as Random Walk and BFS. A key constraint is the sparse distribution of information, where crucial path nodes must be distributed across a wide range of documents to mandate rigorous cross-document information grounding and retrieval. The sampled reasoning paths are further enhanced via approaches including information perturbation and the obfuscation of entities and concepts in $G_d$. For instance, we perform Temporal Obfuscation ("the year end with 5 in late 20th century"), Location Obfuscation ("a country with a population of over 1.4 billion"), Institutional Obfuscation ("a prestigious science and engineering university located in Beijing"), etc.

\paragraph{Question Generation} We identify that high-reasoning-density questions necessitate the ability to synthesize deep and widely scattered contextual information for long-range reasoning. To this end, we categorize deep reasoning in long-context scenarios into following distinct paradigms:

\begin{itemize} 
    \item \textbf{In-depth Reasoning}: It includes \textbf{Multi-hop Reasoning} and \textbf{Causal Analysis}, which focus on tracking intricate logical chains across massive contexts to identify non-obvious dependencies and root causes.
    \item \textbf{Temporal Reasoning}: It requires aggregating discrete quantitative data spread throughout the text to perform precise calculations and model dynamic shifts over time. \item \textbf{Hypothetical Scenario}: It evaluates the ability of counterfactual reasoning by mapping existing logic onto new, speculative frameworks.
\end{itemize}

Based on the source documents and the extracted reasoning paths as ground truth, we employ a strong teacher LLM as the generator to synthesize high-quality multi-hop QA pairs. The generation process integrates the aforementioned paradigms and incorporates question expression obfuscation. We achieved precise control over question complexity by regulating the length of the sampled paths, and ensured the variety of the dataset by using paths derived from various sampling methods.

\paragraph{Quality Control} Ideally, the constructed long-context multi-hop QAs should be well-grounded in supporting documents, with concise answers and high-quality reasoning paths. With QA pairs obtained, we further employ a four-stage quality control pipeline to obtain a high-quality dataset: (1) \textbf{Answer Alignment Check}: teacher LLMs are used as generator, responder, and verifier to ensure consistency between the generated questions and answers; (2) \textbf{Knowledge Grounding Check}: questions that can be answered correctly without the source documents are removed to avoid the potential influence from LLMs' inherent knowledge; (3) \textbf{Complex Answer Filtering}: we retain only QA pairs whose answers are under 20 words to ensure stable and reliable validation; (4) \textbf{Contextual Robustness Check}: samples that are easily perturbed by irrelevant documents are filtered out. More detailed quality control pipeline can be referred at Appendix~\ref{sec:quality_control_pipeline}.

\subsection{Long-Context Reinforcement Learning with Process Advantage Shaping}\label{sec:advantage_reweight}
\begin{figure}[!t]
    \centering
    \includegraphics[width=0.99\textwidth]{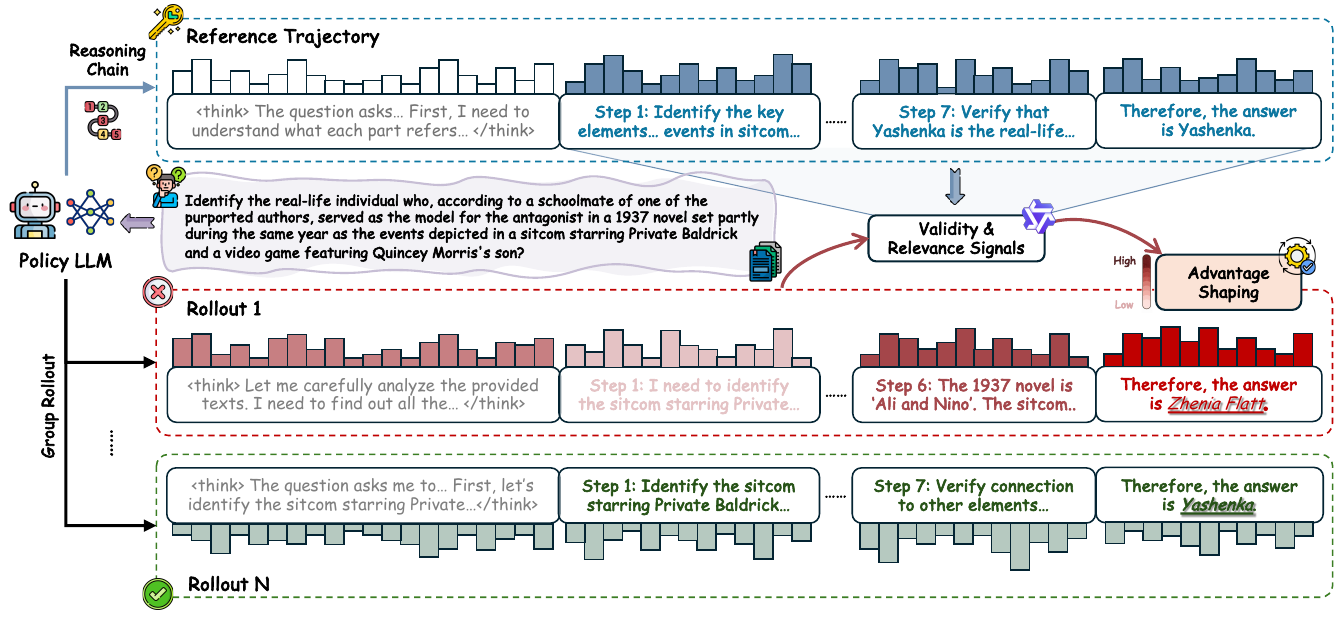}
    \vspace{-0.1cm}
    \caption{Overview of Long-Context Reinforcement Learning with Process Advantage Shaping.}
    \label{fig:algo_pipeline}
    \vspace{-0.25cm}
\end{figure}

Analysis in Section~\ref{sec:analysis_steps_in_negative} unveils the common ``almost-there" phenomenon in existing RLVR methods, highlighting the inherent limitations of relying on sparse, outcome-based rewards tied exclusively to final correctness. Therefore, we introduce \OurMODEL{}, which incorporates process advantage shaping guided by on-policy reference trajectories to facilitate fine-grained credit assignment. The overall framework is illustrated in Figure~\ref{fig:algo_pipeline}.

\paragraph{On-policy Supervision from GT-guided Trajectory} Given a question $Q$ and long context $C$, in addition to regular group sampling rollouts $T$, we perform auxiliary sampling by incorporating the ground-truth reasoning chain into the prompt. This process transforms condensed entity-relation phrases into a natural language trajectory $\tau_p$, ensuring it exhibits reasoning patterns and thought processes consistent with the policy rollouts. Consequently, $\tau_p$ can serve as an on-policy reference supervision signal.
To extract intermediate steps from reasoning trajectories, we craft a simple prompt template that instructs the LLM to structure its response into distinct steps. This template explicitly mandates a step-by-step format to facilitate parsing. The detailed prompt templates can be found in Appendix~\ref{sec:appendix_experimental_details}.

\paragraph{Hybrid Reward Design} Unlike tasks such as Math and Coding in short-context scenarios, which allow for verification via rule-based functions, the long-context reasoning tasks (e.g., open-ended QA) present unique challenges due to inherent answer diversity. Therefore, we adopt a hybrid reward mechanism~\citep{QwenLong_L1} to balance precision and recall. Specifically, we first perform exact string matching between the model prediction $y_{pred}$ and the ground truth $y_{gt}$ as $R_{rule} = \mathbb{I}(y_{pred} = y_{gt})$. To mitigate potential false negatives and handle answer diversity, we additionally employ an LLM-based evaluator to assess the semantic equivalence between $y_{pred}$ and $y_{gt}$ as $R_{LLM} = \text{LLM}_{Judge}(Q,y_{pred}, y_{gt})$, where $Q$ denotes the input question. We then obtain the final reward through maximum selection as follows:
\begin{equation*}
R(Q,y_{pred}, y_{gt}) = \max \left\{ \mathbb{I}(y_{pred} = y_{gt}), \text{LLM}_{Judge}(Q,y_{pred}, y_{gt}) \right\}.
\end{equation*}

\paragraph{Process Advantage Shaping} We posit that the reference GT-guided trajectory $\tau_p$ encapsulates the grounded information scope and essential reasoning steps. This allows us to mitigate the limitation of ``erroneous penalization of correct sub-steps" via stepwise advantage shaping.
Specifically, for group sampling rollouts $T=\{\tau_1,\tau_2,...,\tau_n\}$, we evaluate each sub-step $s_j$ of rollout $\tau_i$ in two dimensions: \textit{validity} and \textit{relevance}. \textit{Validity} is evaluated with an LLM-as-a-Judge, denoted as $\mathbb{I}_{\text{valid}}(s_{i,j})=\text{LLM}_{Judge}(s_j,\tau_p)$. The judge assesses whether $\tau_i$ aligns with the necessary entities and reasoning logic of the reference $\tau_p$. \textit{Relevance} is quantified by semantic similarity $\text{sim}(s_{i,j}, \tau_p)$ and reflects the extent to which the sub-step $s_j$ is semantically aligned with its rollout $\tau_i$. To balance exploration and exploitation, we utilize these signals to perform granular stepwise advantage reweighting specifically on rollouts with negative outcomes, whereas positive rollouts remain intact to foster exploration of diverse reasoning paths. The reweighted stepwise advantage $\hat{A}_{i,j}$ for the $j$-th step $s_{i,j}$ in the $i$-th rollout $\tau_i$ is calculated as:

\begin{equation*}
    \hat{A}_{i,j} = \underbrace{\frac{r_i - \text{mean}(\{r_1,r_2,...,r_n\})}{\text{std}(\{r_1,r_2,...,r_n\})}}_{\text{Group Relative Advantage}} \cdot \underbrace{\left( 1 - \mathbb{I}(\tau_i \in T_f) \cdot \mathbb{I}_{\text{valid}}(s_{i,j}) \cdot \text{sim}(s_{i,j}, \tau_p) \right)}_{\text{Step-wise Reweighting Coefficient}},
\end{equation*}

where $\mathbb{I}(\tau_i \in T_f)$ is an indicator function for negative rollouts, and $\text{sim}(s_{i,j}, \tau_p)$ denotes the semantic similarity between step $s_{i,j}$ and $\tau_p$. Consequently, when a sub-step in a negative rollout is deemed correct, the penalty vanishes ($\hat{A}_{i,j} \to 0$), whereas erroneous steps retain the full negative signal. For Thinking models, since their outputs contain ``thinking trajectories" enclosed by ``\verb|<think>|" and ``\verb|</think>|" tokens, we assign the average advantage of the ``response" part uniformly to each token within this thinking segment.

We opt to mitigate penalties rather than assign positive rewards to avoid optimization ambiguity. Assigning positive advantages within negative rollouts risks incentivizing "plausible but ineffective" reasoning paths that do not lead to correct outcomes. Instead, our approach acts as a conservative credit assignment mechanism: it signals that correct sub-steps were not the cause of failure—preventing them from being ``unlearned".

\paragraph{Overall Training Objective} Following the defined Process Advantage Estimation algorithm, we now formalize the final training objective, which incorporates the Step-wise Re-weighted Advantage ($\hat{A}_{i,j}$). The policy $\pi_{\theta}$ is optimized by maximizing the following objective function:

\begin{equation*}
\scalebox{0.9}{
$J_{\text{GRPO}}(\theta) = \hat{\mathbb{E}}_{(C, Q), \boldsymbol{y}} \left[ \frac{1}{N} \sum_{i=1}^{N} \frac{1}{L_i} \sum_{j=1}^{L_i} f_{\epsilon} \left( \rho_{i,j}(\theta), \hat{A}_{i,j} \right) \right] - \beta \cdot \hat{\mathbb{E}}_{(C, Q)} \left[ \mathbb{D}_{KL}[\pi_{\theta}(\cdot \mid C, Q) \| \pi_{old}(\cdot \mid C, Q)] \right],$
}
\end{equation*}

where $\rho_{i,j}(\theta) = \frac{\pi_{\theta}(s_{i,j} \mid s_{i,<j}, C, Q)}{\pi_{old}(s_{i,j} \mid s_{i,<j}, C, Q)}$ denotes the probability ratio of $j$-th step in $i$-th trajectory.

\section{Experimental Setup}

\paragraph{Data Construction} During the multi-hop QA synthesis pipeline in Section~\ref{sec:QA_synthesis}, we employ KGGen~\citep{KGGen} to extract entities and triplets from plain texts. We sample reasoning paths from 2 to 30 hops to control the difficulties of corresponding questions.

As a result, we construct the raw \OurDataset{} dataset containing 14,577 QA samples with documents up to 288K tokens in length. Prior to training LLMs, we filter the training questions to exclude trivial and insurmountable instances. Specifically, for each question, we perform 8 full-context rollouts with Qwen3-4B-Thinking and retain only those whose empirical success rate lies in the interval [0.25, 0.75]. This yields a filtered RL training dataset of 2,012 QA samples with documents up to 60K tokens. Detailed dataset statistics and data filtering pipeline are provided in Appendix~\ref{sec:dataset_statistics} and~\ref{sec:quality_control_pipeline}.

\paragraph{Training Settings} We conduct experiments on different LLM backbones: (1) Instruct Models: LLaMA3.1-8B-Instruct, Qwen2.5-7B-Instruct, Qwen3-4B-Instruct and Qwen3-30B-A3B-Instruct; (2) Thinking Models: Qwen3-4B-Thinking and Qwen3-30B-A3B-Thinking. We include RLVR as the baseline method implemented by naive GRPO~\citep{GRPO}, and trained on the same dataset. Our RL framework is implemented upon VeRL~\citep{VeRL}. For the RL training, we use the AdamW optimizer with a constant learning rate of 2e-6 and a 5-step linear warmup. The maximum input length is set to 60K tokens, the maximum output length is set to 30K tokens for Thinking models and 10K for Instruct models. We conduct purely on-policy training, use a prompt batch size 128 and set group size $N$ to 8, sampling temperature to 0.7 and top-$p$ to 0.95. As for GT path guided sampling, temperature is set to 0 and top-$p$ is set to 1. We adopt GPT-OSS-120B as the LLM-as-Judge for answer evaluator and provide validity signals in Process Advantage Estimation. We use Qwen3-8B-Embedding to calculate semantic similarity between trajectory steps.

\paragraph{Evaluation Configurations} We evaluate all LLMs on three widely-used and challenging long-context QA benchmarks: (1) FRAMES~\citep{Frames}, which comprises questions requiring 2-15 Wikipedia articles to answer. (2) LongBench V2~\citep{LongBench_v2}, a realistic multi-choice QA benchmark that assess the ability of LLMs to handle long-context problems requiring deep understanding and reasoning, with contexts ranging from 8K to 2M words. (3) Multi-Hop QA: We adopt three subsets including 2WikiMultiHopQA, HotpotQA and MusiQue from Longbench~\citep{LongBench}, which cover 3-5 hop questions with corresponding documents. To contextualize performance, we also evaluate frontier LLMs in different sizes including GPT5-Nano~\citep{GPT5}, Gemini2.5-Flash-Thinking~\citep{Gemini2.5} and GPT-OSS (20B and 120B)~\citep{GPT_OSS}. More detailed evaluation configurations can be found in Appendix~\ref{sec:appendix_experimental_details}.

\section{Experimental Results}

\subsection{Main Results}

\begin{table*}[t]
\centering
\caption{Overall performance of models on long-context QA benchmarks. RLVR is implemented with GRPO~\citep{GRPO}. Results are reported using the \textbf{Avg@4} metric. The top scores for each backbone LLM are \textbf{bolded}. We additionally report four representative reasoning-intensive sub-task performance of LongBench V2, full results are provided in Appendix~\ref{sec:app_longbench_result}.}
\label{tab:main_results}
\setlength\tabcolsep{6pt} 
\renewcommand{\arraystretch}{1.4} 
\fontsize{9pt}{10pt}\selectfont 

\resizebox{\linewidth}{!}{
\begin{tabular}{l|c|cccc|c|ccc|c}
\toprule
\multicolumn{1}{c|}{\multirow{2}{*}{\textbf{Models}}} & \multirow{2}{*}{\textbf{FRAMES}} & \multicolumn{5}{c|}{\textbf{LongBench V2}} & \multicolumn{4}{c}{\textbf{Multi-Hop QA}} \\
\cmidrule(lr){3-7} \cmidrule(lr){8-11}
& & SingleDoc & MultiDoc & Code Repo & Dialogue & \textbf{Overall} & 2Wiki & HotpotQA & MusiQue & \textbf{Avg} \\
\midrule

\multicolumn{11}{c}{\cellcolor{gray!10}\textbf{\textit{Frontier Models}}} \\
\midrule
GPT5-Nano & 73.54 & 44.00 & 39.20 & 50.00 & 46.15 & 43.74 & 89.00 & 82.00 & 61.50 & 77.50 \\
Gemini-2.5-Flash-Thinking & 65.78 & 51.43 & 55.20 & 58.00 & 66.67 & 56.77 & 89.34 & 81.00 & 60.00 & 76.78 \\
GPT-OSS-120B & 72.69 & 44.57 & 43.20 & 53.06 & 61.54 & 47.01 & 89.00 & 82.00 & 66.00 & 79.00 \\
GPT-OSS-20B & 64.44 & 38.51 & 40.80 & 56.00 & 61.54 & 43.37 & 88.50 & 79.00 & 49.25 & 72.25 \\
\midrule

\multicolumn{11}{c}{\cellcolor{gray!10}\textbf{\textit{Instruct Models}}} \\
\midrule
LLaMA3.1-8B-Instruct & 41.84 & 28.86 & 26.20 & 27.00 & 32.05 & 27.93 & 43.62 & 61.75 & 21.00 & 42.12 \\
- RLVR & 55.98 & 31.86 & 29.20 & 29.00 & 32.69 & 28.93 & 68.00 & 69.75 & 46.62 & 61.46 \\
\rowcolor{darkgreen!8}
- \textbf{\OurMODEL} & \textbf{60.62} & 30.14 & 29.80 & 32.00 & 34.62 & \textbf{29.62} & 79.38 & 76.38 & 52.00 & \textbf{69.25} \\
\midrule
Qwen2.5-7B-Instruct & 45.08 & 36.71 & 28.20 & 32.00 & 35.90 & 33.60 & 54.75 & 69.75 & 32.38 & 52.29 \\
- RLVR & 50.97 & 34.43 & 29.00 & 30.00 & 40.38 & 31.01 & 68.25 & 67.75 & 40.25 & 58.75 \\
\rowcolor{darkgreen!8}
- \textbf{\OurMODEL} & \textbf{55.76} & 39.57 & 28.40 & 29.50 & 31.41 & \textbf{33.70} & 80.88 & 80.00 & 51.62 & \textbf{70.83} \\
\midrule
Qwen3-4B-Instruct & 46.81 & 36.00 & 30.60 & 36.50 & 60.26 & 37.28 & 63.25 & 67.62 & 28.75 & 53.21 \\
- RLVR & 60.92 & 39.14 & 37.00 & 49.50 & 60.90 & 42.10 & 82.62 & 75.88 & 49.25 & 69.25 \\
\rowcolor{darkgreen!8}
- \textbf{\OurMODEL} & \textbf{64.90} & 38.14 & 40.60 & 48.00 & 63.46 & \textbf{42.94} & 86.50 & 75.38 & 55.00 & \textbf{72.29} \\
\midrule
Qwen3-30B-A3B-Instruct & 62.96 & 42.43 & 38.40 & 52.00 & 62.82 & 44.43 & 83.88 & 77.62 & 51.50 & 71.00 \\
- RLVR & 66.26 & 50.29 & 50.40 & 42.00 & 56.41 & 47.91 & 86.00 & 76.00 & 57.00 & \textbf{73.00} \\
\rowcolor{darkgreen!8}
- \textbf{\OurMODEL} & \textbf{68.93} & 46.29 & 42.40 & 56.00 & 71.79 & \textbf{49.11} & 84.50 & 77.50 & 55.50 & 72.50 \\
\midrule

\multicolumn{11}{c}{\cellcolor{gray!10}\textbf{\textit{Reasoning Models}}} \\
\midrule
Qwen3-4B-Thinking & 60.84 & 37.00 & 35.60 & 41.50 & 60.26 & 40.46 & 85.50 & 75.12 & 50.25 & 70.29 \\
- RLVR & 62.74 & 37.14 & 40.80 & 46.00 & 66.67 & 41.75 & 83.50 & 71.50 & 53.50 & 69.50 \\
\rowcolor{darkgreen!8}
- \textbf{\OurMODEL} & \textbf{64.75} & 40.29 & 40.00 & 43.00 & 62.82 & \textbf{42.30} & 85.62 & 78.62 & 54.25 & \textbf{72.83} \\
\midrule
Qwen3-30B-A3B-Thinking & 69.66 & 44.00 & 44.00 & 46.00 & 64.10 & 48.31 & 84.31 & 76.94 & 63.44 & 74.90 \\
- RLVR & 69.66 & 40.00 & 46.40 & 50.00 & 51.28 & 44.53 & 84.00 & 77.00 & 64.00 & 75.00 \\
\rowcolor{darkgreen!8}
- \textbf{\OurMODEL} & \textbf{71.84} & 47.43 & 54.40 & 60.00 & 76.92 & \textbf{54.27} & 86.50 & 77.00 & 66.00 & \textbf{76.50} \\
\bottomrule
\end{tabular}}
\end{table*}

\paragraph{Substantial Gains over Various LLMs} 
Table~\ref{tab:main_results} reports the main results of \OurMODEL{} compared to various baselines and frontier LLMs across different model families and scales (4B, 8B, 30B, etc.). We highlight two key observations: (1) Both our synthesized dataset \OurDataset{} and our training method \OurMODEL{} contribute substantial improvements. For example, RLVR yields significant gains over the original LLMs across all benchmarks: Qwen3-4B-Instruct (46.81 vs. 60.92 on FRAMES), Qwen3-4B-Thinking (60.84 vs. 62.74 on FRAMES). These improvements demonstrate the effectiveness of \OurDataset{}. Furthermore, replacing vanilla RLVR with \OurMODEL{} leads to additional gains: Qwen3-4B-Instruct (60.92 vs. 64.90 on FRAMES), Qwen3-4B-Thinking (69.50 vs. 72.83 on Multi-Hop QA); (2) \OurMODEL{} achieves comparable performance to frontier LLMs with much fewer parameter scales. For example, \OurMODEL{} trained on Qwen3-4B (both Instruct and Thinking) attains results on FRAMES and Multi-Hop QA that are comparable with Gemini-2.5-Flash-Thinking and GPT-OSS-20B. In addition, \OurMODEL{} trained on Qwen3-30B-A3B surpasses GPT5-Nano and GPT-OSS-120B on LongBench V2.

\paragraph{Outperform different Training Strategies}
To further investigate how \OurMODEL{} outperforms baseline training approaches, we report the comparison results against SFT, GRPO and DAPO in Table~\ref{tab:training_strategies}. It is evidenced that \OurMODEL{} achieves larger gains than standard RL approaches like GRPO and DAPO~\citep{DAPO} on both Instruct and Thinking models. Given that \OurMODEL{} exploits ground-truth reasoning chains, we also include a SFT baseline: we distill reasoning trajectories from a teacher LLM (Gemini2.5-Pro) under the guidance of GT reasoning chains, and then supervised fine-tune the student LLM. However, SFT lags behind the RL-based methods, which exploit GT reasoning chains in an on-policy manner, highlighting the advantage of RL-style process optimization over offline distillation. Furthermore, the diminished performance on LongBench V2 suggests that SFT suffers from limited generalization to out-of-domain long-context tasks.

\begin{wrapfigure}[16]{r}{0.60\textwidth}
    \centering
    \vspace{-0.5cm}
    {\includegraphics[width=0.60\textwidth]{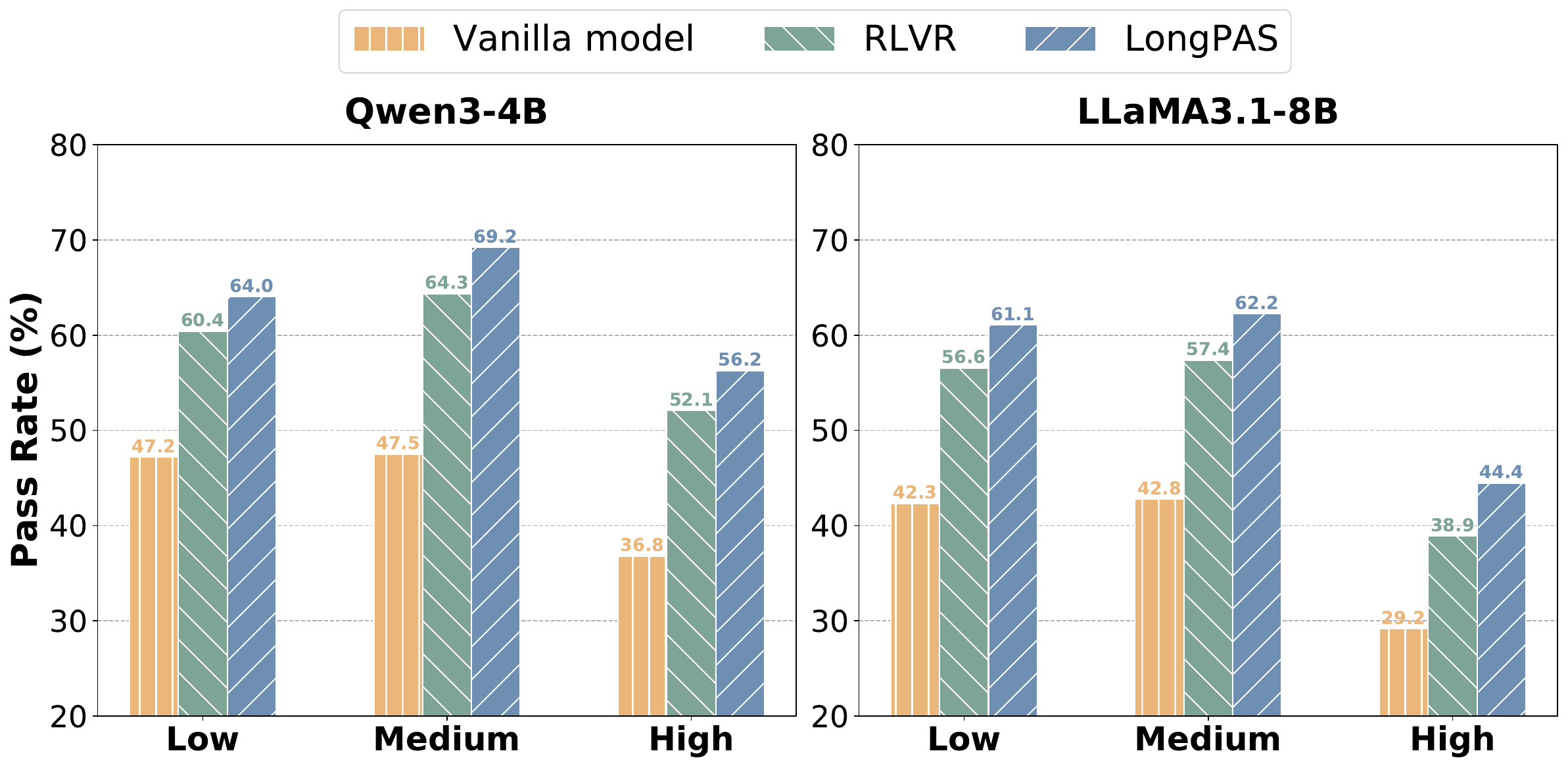}}
    \caption{Performance of \OurMODEL{} on FRAMES with different hop numbers. Questions are categorized into three complexities according to hop numbers: Low ($\leq$3), Medium (4-6) and High ($\geq$7).}
    \label{fig:frames_hop_analysis}
\end{wrapfigure}

\paragraph{Consistent Improvements at Increasing Reasoning Depth}
To further investigate the effectiveness of our proposed \OurMODEL{} in different long-context reasoning difficulties,
we categorize questions in FRAMES benchmark into three groups—Low ($\leq$3), Medium (4–6), and High ($\geq$7)—and report the performance of \OurMODEL{} on different reasoning depths. As shown in Figure~\ref{fig:frames_hop_analysis}, across all settings, both RLVR and \OurMODEL{} improve substantially over the vanilla models; however, the advantage of \OurMODEL{} becomes particularly pronounced as the reasoning complexity increases. For Qwen3‑4B, while the vanilla model’s performance drops markedly from 47.2\% on low-hop questions to 36.8\% on high-hop ones, \OurMODEL{} maintains a much stronger performance, achieving 56.2\% on high-complexity questions—an improvement of nearly 20 percentage points over the baseline. This gain is larger than that of RLVR (52.1\%), indicating that \OurMODEL{} is especially effective at stabilizing and enhancing the model’s ability to handle long reasoning chains.

\begin{table*}[t]
\centering
\caption{Pass@1 Performance Comparison of different training strategies on long-context QA benchmarks. For SFT training, we distill high-quality QAs with reasoning trajectories through DeepSeek-V3 and DeepSeek-R1 (for Instruct and Thinking model individually) under the guidance of GT reasoning chains.}
\label{tab:training_strategies}
\setlength\tabcolsep{6pt} 
\renewcommand{\arraystretch}{1.3} 
\fontsize{9pt}{10pt}\selectfont 

\resizebox{\linewidth}{!}{
\begin{tabular}{l|c|cccc|c|ccc|c}
\toprule
\multicolumn{1}{c|}{\multirow{2}{*}{\textbf{Models}}} & \multirow{2}{*}{\textbf{FRAMES}} & \multicolumn{5}{c|}{\textbf{LongBench V2}} & \multicolumn{4}{c}{\textbf{Multi-Hop QA}} \\
\cmidrule(lr){3-7} \cmidrule(lr){8-11}
& & SingleDoc & MultiDoc & Code Repo & Dialogue & \textbf{Overall} & 2Wiki & HotpotQA & MusiQue & \textbf{Avg} \\
\midrule

Qwen3-4B-Instruct & 46.81 & 36.00 & 30.60 & 36.50 & 60.26 & 37.28 & 63.25 & 67.62 & 28.75 & 53.21 \\
\midrule
- SFT & 55.10 & 28.57 & 30.40 & 26.00 & 41.03 & 31.41 & 79.50 & 74.00 & 49.00 & 67.50 \\
- GRPO & 60.92 & 39.14 & 37.00 & 49.50 & 60.90 & 42.10 & 82.62 & 75.88 & 49.25 & 69.25 \\
- DAPO & 62.11 & 37.57 & 36.60 & 43.00 & 58.33 & 40.26 & 85.00 & 76.00 & 51.12 & 70.71 \\
\rowcolor{darkgreen!8}
- \textbf{\OurMODEL{} (Ours)} & \textbf{64.90} & 38.14 & 40.60 & 48.00 & 63.46 & \textbf{42.94} & 86.50 & 75.38 & 55.00 & \textbf{72.29} \\
\midrule

Qwen3-4B-Thinking & 60.84 & 37.00 & 35.60 & 41.50 & 60.26 & 40.46 & 85.50 & 75.12 & 50.25 & 70.29 \\
\midrule
- SFT & 59.95 & 37.14 & 31.20 & 36.00 & 53.85 & 35.19 & 84.50 & 72.50 & 49.50 & 68.83 \\
- GRPO & 62.74 & 37.14 & 40.80 & 46.00 & 66.67 & 41.75 & 83.50 & 71.50 & 53.50 & 69.50 \\
- DAPO & 62.01 & 31.43 & 36.80 & 42.00 & 53.85 & 35.98 & 86.00 & 77.00 & 53.00 & 72.00 \\
\rowcolor{darkgreen!8}
- \textbf{\OurMODEL{} (Ours)} & \textbf{64.75} & 40.29 & 40.00 & 43.00 & 62.82 & \textbf{42.30} & 85.62 & 78.62 & 54.25 & \textbf{72.83} \\

\bottomrule
\end{tabular}}
\end{table*}

\subsection{Ablation Study}

In this section, we conduct ablation studies to investigate the key components in \OurMODEL{} that achieve long-context in-depth reasoning, including (1) the training data length; (2) process signals in advantage estimation; and (3) the role of on-policy trajectory supervision.

\begin{wrapfigure}[14]{r}{0.60\textwidth}
    \centering
    \vspace{-0.5cm}
    {\includegraphics[width=0.60\textwidth]{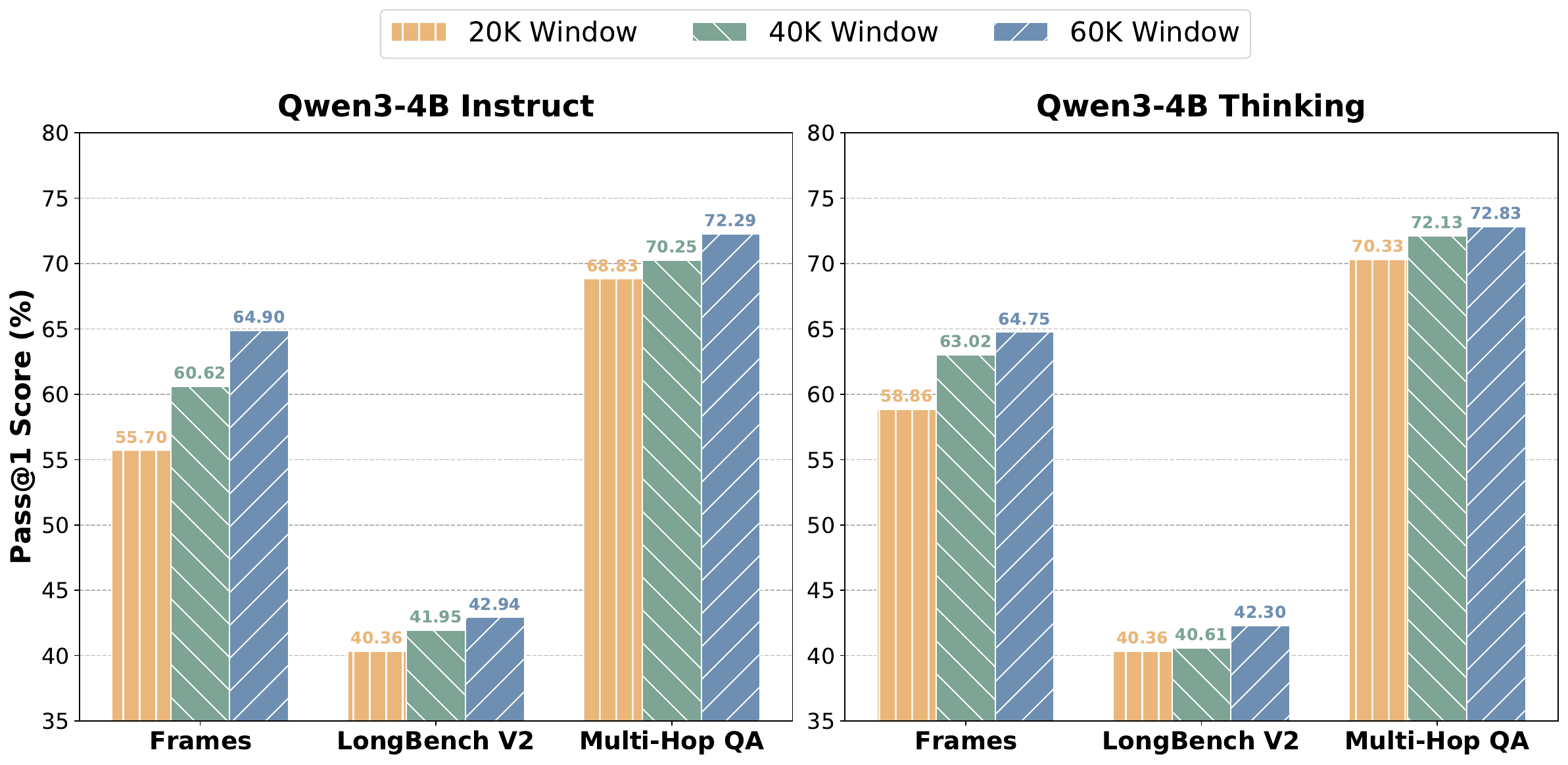}}
    \caption{Performance of \OurMODEL{} trained on different context windows (20K, 40K \& 60K).}
    \label{fig:training_data_length_bar}
\end{wrapfigure}

\paragraph{Training Data Length} We investigate the impact of training data length on the performance of \OurMODEL{}. Specifically, we train Qwen3-4B with varying maximum input lengths—20K, 40K, and 60K tokens—aligned with different context window sizes. As shown in Figure~\ref{fig:training_data_length_bar}, the performance of the \OurMODEL{} model, across both Qwen3-4B-Instruct and Qwen3-4B-Thinking variants, exhibits a monotonic increase with the expansion of the maximum input length during training, utilizing 20K, 40K, and 60K token context windows. This observation underscores the critical role of long-sequence training in improving the model's long-context grounding and reasoning. Specifically, the FRAMES and Multi-Hop QA tasks register the most substantial performance gains. This indicates that training with extended context windows significantly enhances the model's capability for complex information grounding and long-range dependency modeling, which are prerequisites for multi-step reasoning. While performance gains on LongBench V2 benchmark are less pronounced, the consistent positive correlation confirms that greater exposure to longer contexts during training systematically elevates the model's overall proficiency across diverse long-context tasks.

\paragraph{Process Signal of Advantage Shaping} In Section~\ref{sec:advantage_reweight}, we estimate the stepwise advantage shaping signals along two dimensions: \textit{Validity} and \textit{Relevance}. We perform an ablation by removing the \textit{Validity} signal $\mathbb{I}_{\text{valid}}(s_{i,j})=\text{LLM}_{Judge}(s_j,\tau_p)$, and retaining only \textit{Relevance} as the indicator of the necessity of sub-steps. As reported in Table~\ref{tab:ablation_study}, removing the \textit{Validity} signal causes a clear performance decline across all benchmarks, especially on long-context multi-hop reasoning tasks. On FRAMES, performance decreases from 64.90 to 62.68 (-2.28\%), and on LongBench V2 it drops from 42.94 to 41.15 (-1.79\%). We also evaluate an ablation that removes \textit{Relevance} signal while keeping only \textit{Validity} signal, which binarizes the continuous reweighting coefficient. A similar performance degradation is observed: -1.73\% on FRAMES and -2.04\% on Multi-Hop QA.
These reductions indicate that both the \textit{Validity} and \textit{Relevance} signal play an essential role in filtering out incorrect or noisy intermediate steps, thereby ensuring that advantage shaping emphasizes only logically sound reasoning transitions.

\paragraph{On-policy Trajectory Supervision} During Process Advantage Shaping, we sample GT-guided reference trajectory via an on-policy strategy. To explore the role of on-policy supervision in Process Advantage Shaping, we conduct an ablation comparison by replacing it with off-policy supervision from teacher LLMs. Specifically, we sample GT-guided reference trajectories $\tau_d$ from Gemini2.5-Pro by incorporating the ground-truth reasoning chain into the prompt. As shown in Table~\ref{tab:ablation_study}, we can also observe a performance drop across all benchmarks, especially on multi-hop reasoning benchmarks (-2.01\% on FRAMES). We conclude this phenomenon for the following reasons: (1) Strategy-Data Distribution Mismatch: While $\tau_p$ provides a stable, efficient process advantage signal (expecially \textit{Relevance} signal) by aligning with the current policy's capability range $P_{\pi}(\tau)$, the off-policy $\tau_d$ comes from a powerful Teacher LLM ($P_{\text{Teacher}}(\tau)$) that is out-of-distribution for the student model. This causes the training model to struggle with accurately estimating the advantage of paths it rarely explores, leading to unstable optimization; (2) Teacher LLM's reasoning steps may be too complex or rely on knowledge organization the training model $\pi$ has not yet learned, and because $\tau_d$ is static, it cannot dynamically adapt to the policy's learning progress like $\tau_p$.

\begin{table*}[t]
\centering
\caption{Ablation Study of \OurMODEL{} on process signals (\textit{Validity} and \textit{Relevance}) and on-policy supervision over long-context QA benchmarks.}
\label{tab:ablation_study}
\setlength\tabcolsep{6pt}
\renewcommand{\arraystretch}{1.3}
\fontsize{9pt}{10pt}\selectfont

\resizebox{\linewidth}{!}{
\begin{tabular}{l|c|cccc|c|ccc|c}
\toprule
\multicolumn{1}{c|}{\multirow{2}{*}{\textbf{Models}}} & \multirow{2}{*}{\textbf{FRAMES}} & \multicolumn{5}{c|}{\textbf{LongBench V2}} & \multicolumn{4}{c}{\textbf{Multi-Hop QA}} \\
\cmidrule(lr){3-7} \cmidrule(lr){8-11}
& & SingleDoc & MultiDoc & Code Repo & Dialogue & \textbf{Overall} & 2Wiki & HotpotQA & MusiQue & \textbf{Avg} \\
\midrule

Base Model & 46.81 & 36.00 & 30.60 & 36.50 & 60.26 & 37.28 & 63.25 & 67.62 & 28.75 & 53.21 \\
\midrule
\rowcolor{darkgreen!8}
\textbf{\OurMODEL} & \textbf{64.90} & 38.14 & 40.60 & 48.00 & 63.46 & \textbf{42.94} & 86.50 & 75.38 & 55.00 & \textbf{72.29} \\
- \textit{w/o} \textit{Validity} Signal & 62.68 & 37.71 & 33.60 & 46.00 & 61.54 & 41.15 & 86.00 & 72.50 & 46.50 & 68.33 \\
- \textit{w/o} \textit{Relevance} Signal & 63.17 & 40.29 & 35.20 & 49.00 & 62.18 & 41.95 & 85.00 & 74.75 & 51.00 & 70.25 \\
- \textit{w/o} On-policy Supervision & 62.89 & 40.57 & 38.40 & 47.50 & 65.38 & 42.45 & 85.50 & 76.62 & 52.88 & 71.67 \\

\bottomrule
\end{tabular}}
\end{table*}

\subsection{Analysis}

\begin{figure}[!t]
    \centering
    \includegraphics[width=0.99\textwidth]{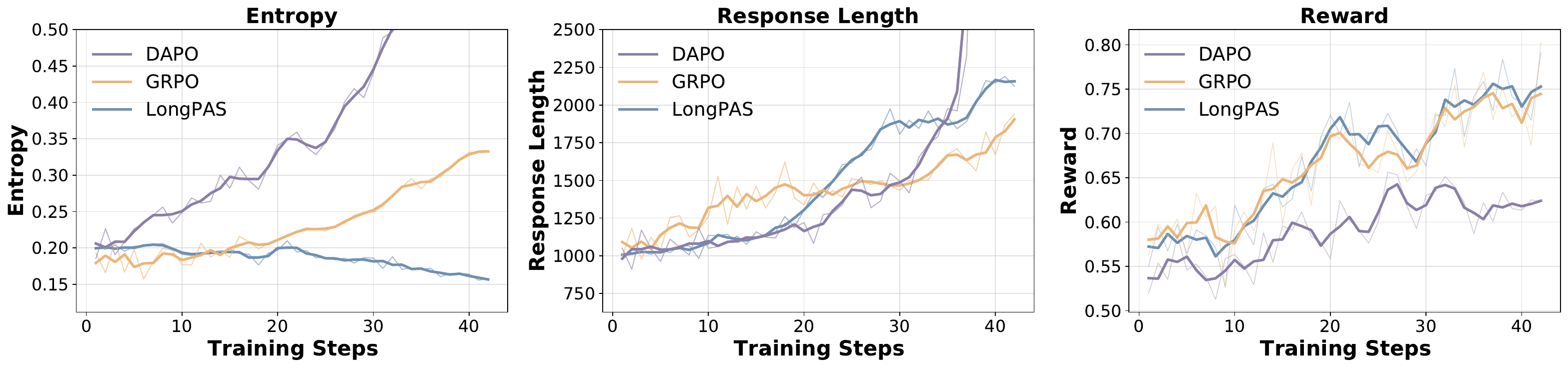}
    \vspace{-0.1cm}
    \caption{Training dynamics of \OurMODEL{} on Qwen3-4B model compared with baseline algorithms. \textbf{Left}: Generation Entropy; \textbf{Middle}: Response Length; \textbf{Right}: Training Reward.}
    \label{fig:training_dynamics}
    \vspace{-0.25cm}
\end{figure}

\paragraph{Training Dynamics}
To understand how \OurMODEL{} achieves strong performance, we report its training dynamics in Figure~\ref{fig:training_dynamics}, including generation entropy, average response length and training reward on \OurDataset{}. 
We can observe that \OurMODEL{} produces markedly more stable training dynamics than GRPO and DAPO on \OurDataset{}.
Regarding the entropy curves, baseline algorithms penalize every step in an ``almost‑there" but ultimately wrong trajectory, the resulting miscredit forces the model into over-exploration, driving entropy to unstable levels. In contrast, \OurMODEL{} offers fine‑grained credit assignment, preserving correct intermediate steps and keeping entropy stable throughout training.
Regarding the response length, \OurMODEL{} keeps responses concise in the early and mid stages and increase length only when it improves performance, effectively regularizing verbosity and preventing exploitation through long, low‑quality reasoning chains. In contrast, DAPO exhibits steadily increasing response lengths and ultimately collapses, degenerating into random, meaningless token generation. Considering training reward, while \OurMODEL{} initially lagging slightly behind GRPO in the early steps, it exhibits strong long-term optimization capabilities. \OurMODEL{}'s reward curve maintains a consistent upward trend, successfully converging with and ultimately surpassing GRPO in the later stages of training. It confirms \OurMODEL{}'s superior performance in maximizing the reward throughout the entire training process.

\paragraph{Triplet Coverage Dynamics}

\begin{wrapfigure}[16]{r}{0.5\textwidth}
    \centering
    \vspace{-0.6cm}
    {\includegraphics[width=0.48\textwidth]{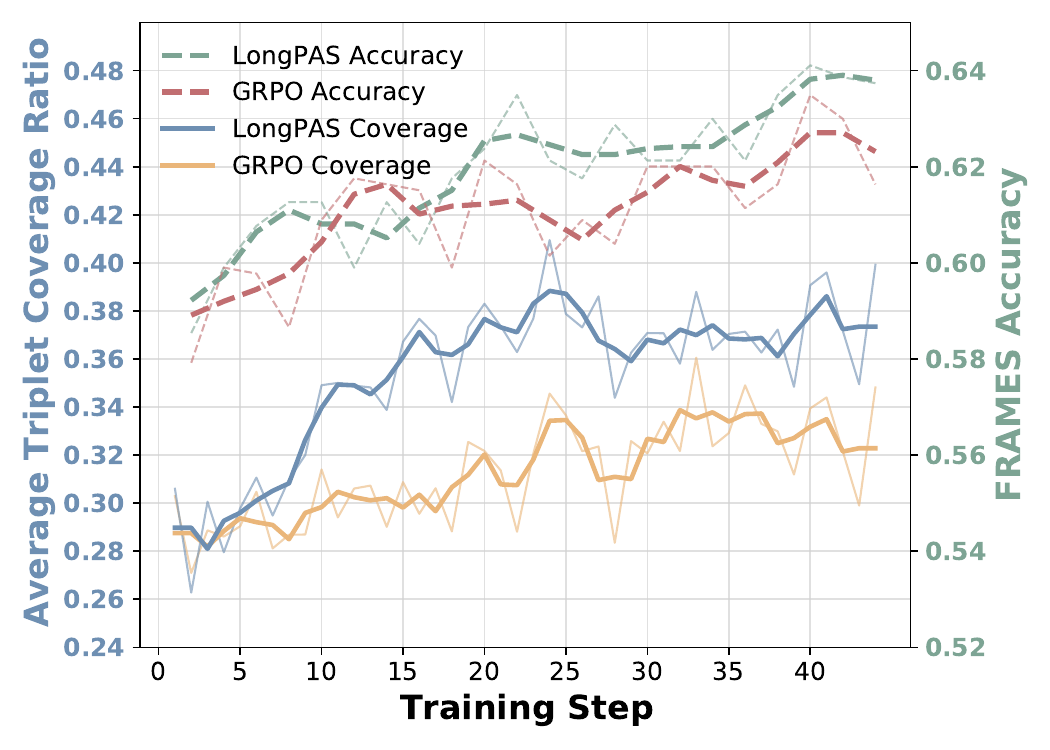}}
    \vspace{-0.2cm}
    \caption{Average Triplet Coverage Ratio (\%) on the training data and FRAMES Accuracy dynamics with Qwen-4B model.}
    \label{fig:triplet_coverage_dynamics}
\end{wrapfigure}

To further investigate the mechanism of \OurMODEL{} in mitigating the common wrong credit assignment confronting ``almost-there" samples, we analyze the dynamics comparison between average Triplet Coverage Ratio and FRAMES Accuracy during training period (Entity Coverage Dynamics are shown in Appendix~\ref{app:entity_coverage}). As shown in Figure~\ref{fig:triplet_coverage_dynamics}, while both methods start at a similar baseline, \OurMODEL{} quickly establishes a dominant lead in triplet coverage, stabilizing at 0.37–0.38, significantly outperforming GRPO. This sustained advantage indicates that \OurMODEL{} more effectively strengthens the model's reasoning capacity by ensuring a higher density of correct reasoning steps. The synchronized growth between coverage and accuracy confirms that grounding the reasoning process in structured evidence yields more precise step-level credit assignment. Ultimately, by rewarding the successful identification of key triplets, \OurMODEL{} effectively mitigates the "almost-there" problem, leading to a more logical, evidence-based reasoning trajectory and superior overall performance.

\section{Related Works}

\subsection{Long-Context Data Synthesis}
Most existing approaches to long-context data synthesis primarily focused on superficially extending input length by augmenting existing datasets with filler content. Specifically, methods such as those by \cite{ALR2,LC_LLM_self_improve,LC_LLM_MHQA} involved padding questions with additional irrelevant documents, for example, by adding unrelated passages to MusiQue~\citep{MusiQue} or using document-filling on HotpotQA~\citep{HotpotQA} and SQuAD~\citep{SQuAD}. Although these strategies effectively increase the sheer volume of context, they are fundamentally limited as they fail to generate the high-quality, challenging training data necessary to develop models capable of deep reasoning.

In contrast, recent advancements in long-context instruction data generation introduce several distinct approaches aimed at creating more complex and faithful contexts. LongFaith~\citep{LongFaith} focuses on synthesizing faithful reasoning datasets by incorporating ground truth and citation-based prompting. To achieve diversity and realism, Wildlong~\citep{WildLong} uses a multi-pronged approach involving meta-information extraction, graph-based modeling, and adaptive generation. Other methods concentrate on generating the context itself: \citep{short_to_long} utilizes off-the-shelf LLMs to create extended background contexts for existing pairs, while Quest~\citep{Quest} employs a query-centric method to aggregate diverse, relevant documents. Re3Syn~\citep{Re3Syn} provides a structured pipeline using semantic retrieval, dependency recognition, and a reorder algorithm to transform short documents into coherent long-context data. SPELL~\citep{SPELL} proposes a multi-role self-play reinforcement learning framework to construct high-quality long-context QA data at scale.

\subsection{Long-Context Reasoning}
While high-quality, human-like chains of thought (CoT) have significantly driven recent advances in LLM reasoning, most existing studies have primarily focused on short-context tasks, such as mathematics~\citep{Deepseek_r1, rStar_Math, L1} and code~\citep{rStar_Coder, open_code_reasoning}. Consequently, the exploration of advanced long-context reasoning patterns remains notably limited. Although previous works successfully extended model context windows during pre- and mid-training stages~\citep{Qwen3, Gemini2.5, Kimi_K2} using methods like RoPE~\citep{RoPE, Yarn}, the resulting models often fail to exhibit strong reasoning capabilities or reliably utilize information across the extended context.

Yet, there are limited researches explored the role of post-training in unlocking LLMs' abilities of robust long-context reasoning. Long-context SFT~\citep{Long_SFT} and DPO~\citep{Long_DPO} tend to introduce non-generalizable bias and noise. More recent efforts like QwenLong-L1~\citep{QwenLong_L1}, SoLoPO~\citep{SoLoPO} and E-GRPO~\citep{EGRPO} employ standard Reinforcement Learning with Verifiable Rewards (RLVR) to incentivize long-context reasoning. However, these existing works predominantly concentrate on optimizing RL training strategies themselves, often overlooking the crucial role of synthesizing high-quality long-context reasoning data. While methods like LoongRL~\citep{LoongRL} attempt to address the data gap by proposing a data-driven RL approach—transforming standard multi-hop questions into high-difficulty long-context tasks through the inclusion of distracting documents—they remain susceptible to an inherent RL challenge: the ubiquitous ``almost-there" phenomenon during training. This issue frequently discards valuable learning signals, which ultimately leads to unstable training dynamics when applied to complex reasoning models.

\section{Conclusion}

In this work, we systematically investigate the ``almost-there" phenomenon within long-context RLVR, highlighting how sparse, outcome-based rewards often overlook pivotal learning signals in partially correct trajectories. To address this, we propose a KG-driven synthesis framework to construct \OurDataset{}, a high-quality multi-hop QA dataset featuring explicit reasoning chains. Building on this, we introduce \OurMODEL{}, which enables fine-grained credit assignment via process advantage shaping. Experimental results demonstrate that our approach significantly enhances the long-context reasoning capabilities of LLMs, consistently outperforming established baselines and allowing smaller models to achieve parity with frontier LLMs. Further analysis reveals that \OurMODEL{} not only stabilizes reinforcement learning but also fosters robust information grounding and complex logical integration. This paradigm—integrating structured data synthesis with granular process supervision—establishes a scalable path toward next-generation reasoning agents capable of navigating increasingly intricate, large-scale information environments.

\section*{Limitations}

This paper aims to incentivize in-depth reasoning of long-context LLMs on long-range multi-hop QA tasks, and offers a recipe combining a KG-guided multi-hop QA synthesis framework with a process advantage shaping strategy. While it makes substantial progress in long-context reasoning, several limitations remain:

\paragraph{Data Domain and Source} The synthesis framework currently relies primarily on Wikipedia as the underlying knowledge source. While \OurDataset{} already yields substantial improvements and effectively incentivizes deep long-context reasoning patterns, incorporating long documents from domains such as law, finance, and medicine could introduce richer stylistic and structural diversity, potentially improving robustness and transferability to real-world long-context reasoning tasks.

\paragraph{Coupling of Synthesis Framework and Process Supervision} A potential limitation is the perceived coupling between LONGPAS and the KG-driven synthesis framework, as the algorithm currently utilizes the specific reasoning chains generated during synthesis as the primary source of process signals. While this might appear to restrict the scalability of LONGPAS to datasets where explicit ground-truth reasoning paths are available , we argue that the core contribution of LONGPAS lies in its generalizable mechanism for Process Advantage Shaping. Specifically, the method provides a robust framework for effectively leveraging any form of auxiliary supervision—whether they are KG-derived reasoning chains or standard reasoning trajectories distilled from teacher LLMs—to stabilize long-context RL training. By transforming these signals into fine-grained credit assignment, LONGPAS addresses the fundamental "almost-there" bottleneck in a way that aligns with contemporary efforts~\citep{SRL, CARFT} to integrate SFT-level supervision into the RL stage. Thus, rather than being a restricted implementation, our approach offers a novel and versatile perspective on mitigating reward sparsity in complex, long-context scenarios.


\paragraph{Sophistication of the Reward Model} Our current implementation uses a hybrid reward function that combines simple rule-based checks with an LLM-as-a-judge to balance precision and recall. This works well for tasks with clear correctness criteria, such as factoid multi-hop questions, but may be less effective for open-ended or subjective tasks where correctness is multifaceted. A promising direction is to develop more advanced, rubric-based LLM reward models that score responses along dimensions such as logical rigor, citation accuracy, and coherence, enabling our framework to better handle complex, agentic scenarios.

\bibliography{iclr2025_conference}
\bibliographystyle{iclr2025_conference}

\clearpage

\appendix

\section{Synthetic Data Details}

\subsection{Dataset Statistics}\label{sec:dataset_statistics}

Table~\ref{tab:dataset_statistics_table} presents the detailed data statistics for \OurDataset{}, providing detailed information on the Raw Synthesis Dataset and the Filtered RL Training Dataset for better understanding. Figure~\ref{fig:category_distribution} display the distribution of QA categories in raw \OurDataset{}. For readability, only categories with top 40 frequency are displayed. In Figure~\ref{fig:question_word_cloud}, we show the word cloud of questions in raw \OurDataset{}.

\begin{figure*}[!h]
    \centering
    \subfloat[Token Length Distribution]{\includegraphics[width=0.49\textwidth]{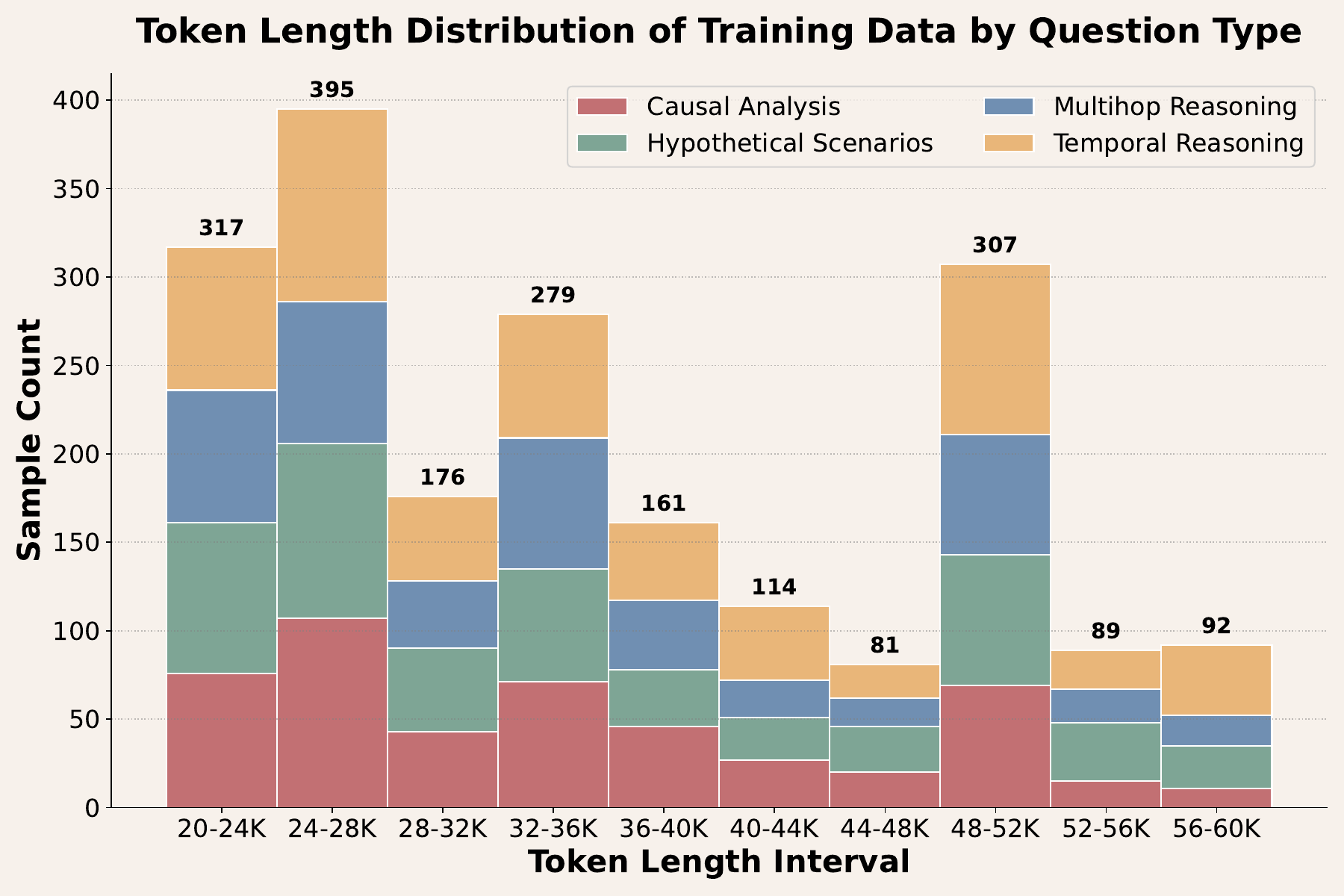}
    \label{token_length_distribution}}  
    \hfill
    \subfloat[Hop Number Distribution]{\includegraphics[width=0.49\textwidth]{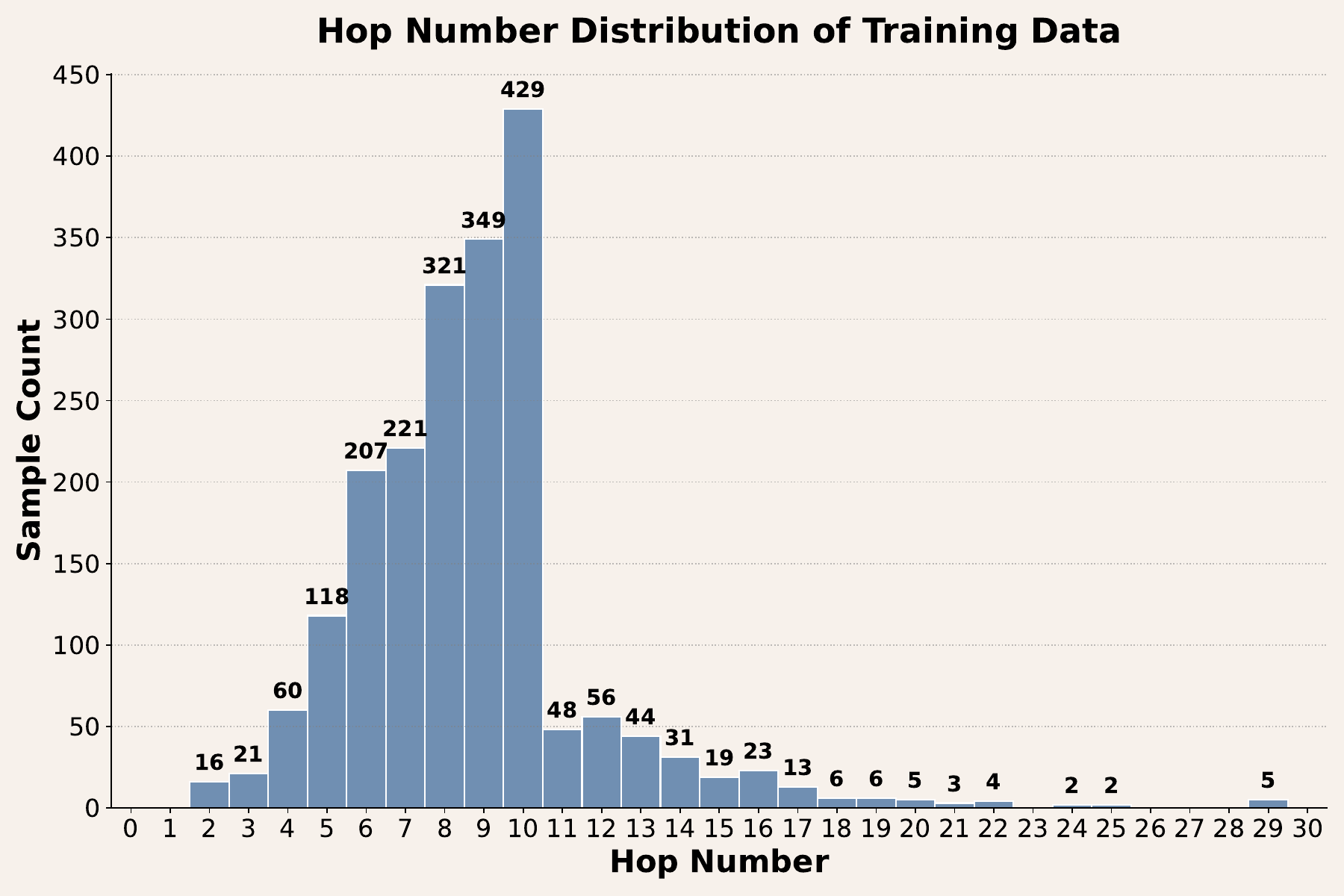}
    \label{hop_number_distribution}}  
    \caption{Distributions of token length (a) and hop numbers (b) in \OurDataset{}. Texts are tokenized by Qwen3-4B-2507 tokenizer.}
    \label{fig:dataset_statistics_bar}
\end{figure*}

Furthermore, we summarize the data distribution of \OurDataset{} in Figure~\ref{fig:dataset_statistics_bar}. Figure~\ref{token_length_distribution} illustrates the token‑length distribution of our training samples, which spans from 20K to 60K. Figure~\ref{hop_number_distribution} shows the hop‑number distribution, revealing that questions produced by our KG‑based QA synthesis framework are highly challenging, with reasoning hops ranging widely from 2 to 30.

\begin{table}[!h]
  \centering
  \begin{minipage}[b]{0.58\linewidth}
    \centering
    \includegraphics[width=\linewidth]{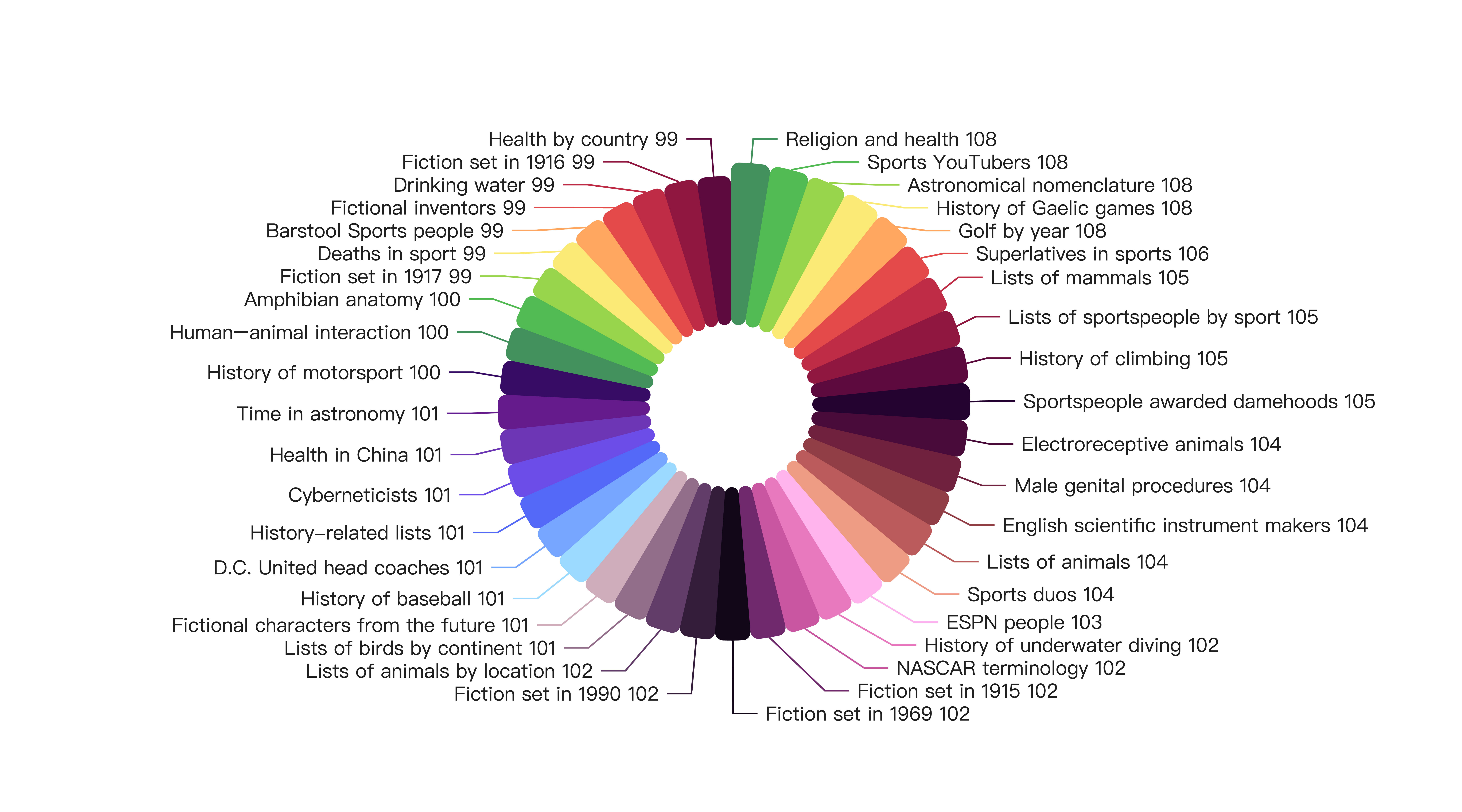}
    \captionof{figure}{Category Distribution of \OurDataset{}}
    \label{fig:category_distribution}
    
    \vspace{0.2cm}
    
    \includegraphics[width=\linewidth]{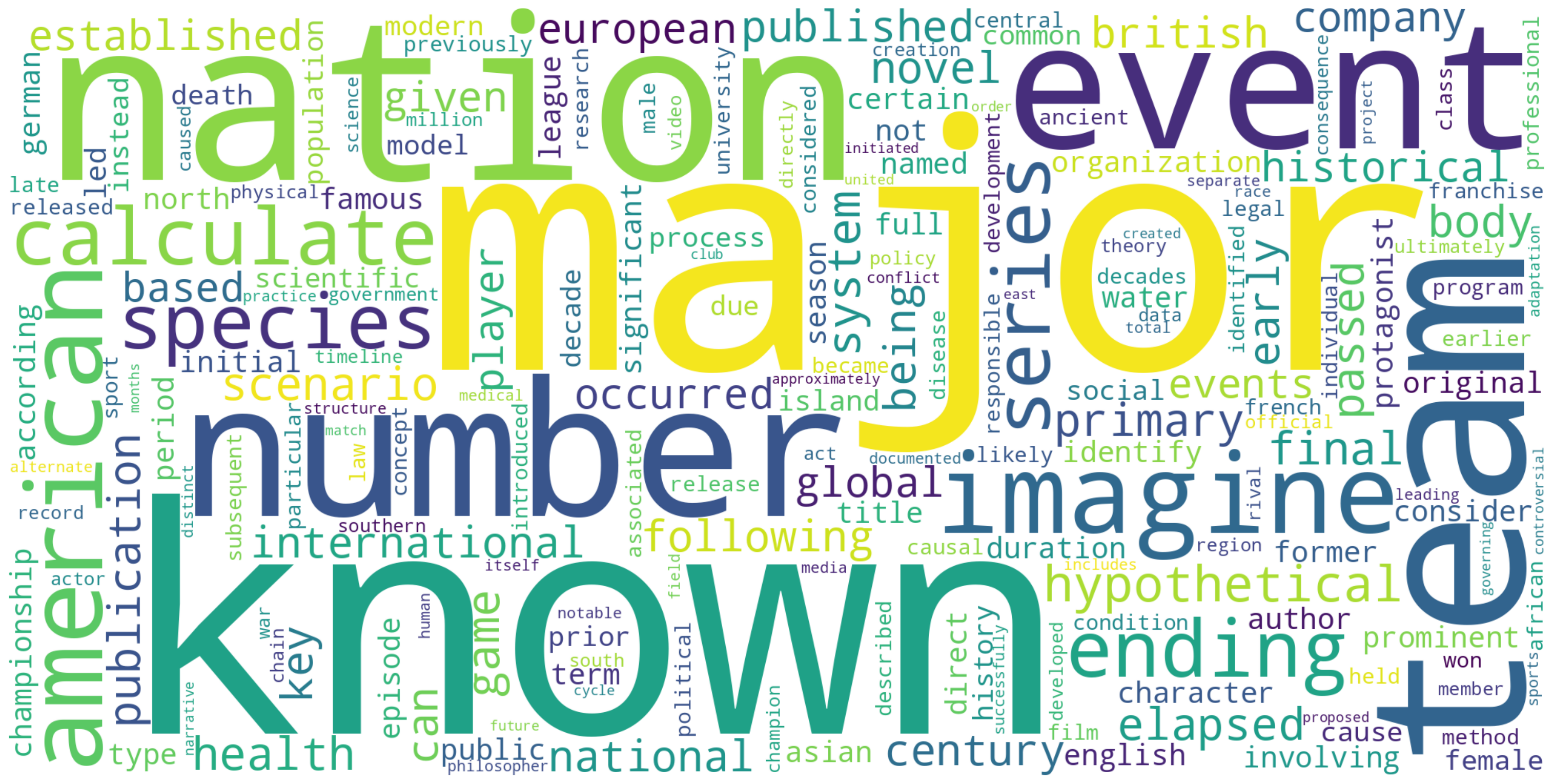}
    \captionof{figure}{Question Word Cloud of \OurDataset{}}
    \label{fig:question_word_cloud}
  \end{minipage}
  \hfill
  \begin{minipage}[b]{0.40\linewidth}
    \centering
    \renewcommand{\arraystretch}{1.26}
    \resizebox{\linewidth}{!}{
    \small
    \begin{tabular}{l|c}\toprule
        \textbf{Property} & \textbf{Number}\\
        \midrule
        \multicolumn{1}{l}{\textbf{\textit{Raw Synthesis Dataset}}} \\
        \# of Topic Categories & 158 \\
        \# of Total QA Pairs & 14,577 \\
        \# of Multi-hop Reasoning Question & 4,415 \\
        \# of Temporal Reasoning Question & 4,345 \\
        \# of Hypothetical Scenarios Question & 3,137 \\
        \# of Causal Analysis Question & 2,680 \\
        \# of Average Hop Number & 8.78 \\
        \# of Minimum Token Number & 10,137 \\
        \# of Maximum Token Number & 288,515 \\
        \# of Average Token Number & 44,703 \\
        \midrule
        \multicolumn{1}{l}{\textbf{\textit{Filtered RL Training Dataset}}} \\
        \# of Total QA Pairs & 2,012 \\
        \# of Multi-hop Reasoning Question & 447 \\
        \# of Temporal Reasoning Question & 572 \\
        \# of Hypothetical Scenarios Question & 508 \\
        \# of Causal Analysis Question & 485 \\
        \# of Average Hop Number & 8.78 \\
        \# of Minimum Token Number & 20,000 \\
        \# of Maximum Token Number & 59,888 \\
        \# of Average Token Number & 35,876 \\
        \bottomrule
    \end{tabular}}
    \caption{Data Statistics of \OurDataset{}}
    \label{tab:dataset_statistics_table}
  \end{minipage}
\end{table}

\subsection{Detailed Quality Control Pipeline}\label{sec:quality_control_pipeline}
Ideally, the constructed long-context multi-hop QAs should be well-grounded in supporting documents, with concise answer and high-quality reasoning paths. Thus, we applied a four‑stage filtering pipeline to ensure the final \OurDataset{} is of high quality:

\paragraph{Answer Alignment Check:} During question generation, we use Gemini‑2.5‑Pro~\citep{Gemini2.5} to produce QA pairs based on task‑specific prompts, documents, and reasoning paths. DeepSeek-V3‑2508~\citep{deepseek_v3} is then used to answer the generated questions, and GPT‑OSS‑120B~\citep{GPT_OSS} serves as a verifier to assess whether the two answers are consistent. Samples with misaligned answers are removed.

\paragraph{Knowledge Grounding Check:} To reduce the potential bias from internal inherent knowledge, we temporarily remove the source documents and check whether the model can still answer correctly. Samples that remain answerable without the documents are filtered out to ensure the dataset genuinely tests contextual reasoning.

\paragraph{Complex Answer Filtering:} QA pairs whose answers exceed 20 words are discarded, as overly complex answers are unstable and difficult to verify reliably.

\paragraph{Contextual Robustness Check:} We augment each context with irrelevant documents and re‑evaluate the model’s answer. Samples whose answer accuracy (pass@k) drops to zero are removed, ensuring that each question–answer pair is robust rather than brittle under context perturbations.

\section{More Experimental Details}\label{sec:appendix_experimental_details}

\paragraph{Benchmark Details} In our experiments, we evaluate the long-context capabilities of our model and baselines on several widely-used benchmarks:

\begin{itemize}
    \item  \textbf{FRAMES}~\citep{Frames}: FRAMES is a comprehensive evaluation dataset designed to test the capabilities of LLMs across factuality, retrieval accuracy, and reasoning. It contains 824 complex, multi-hop reasoning questions related to a diverse set of topics from Wikipedia, involving subjects such as history, sports, science, animals, health, etc. Each question is labeled based on five reasoning types (i.e, Numerical Reasoning, Tabular Reasoning, Multiple Constraints, Temporal Reasoning, and Post-Processing), which requires 2-15 Wikipedia articles to answer. To align with our long-context setting, we equip each question with corresponding wikipedia pages to minimize the effect of information retrieval.
    \item  \textbf{LongBench V2}~\citep{LongBench_v2}: LongBench V2 is a benchmark designed to evaluate LLMs’ ability to tackle long‑context tasks that require deep comprehension and multi‑step reasoning across real‑world scenarios. It comprises 503 challenging multiple‑choice questions with context lengths ranging from 8K to 2M words, covering six key task categories: single‑document QA, multi‑document QA, long in‑context learning, long‑dialogue understanding, code repository comprehension, and long structured‑data understanding.
    \item  \textbf{Multi-Hop QA}~\citep{LongBench}: Multi-Hop QA consist of three subsets including 2WikiMultiHopQA~\citep{2WikiMultiHopQA}, HotpotQA~\citep{HotpotQA} and MusiQue~\citep{MusiQue} that are adopted from LongBench~\citep{LongBench}. HotpotQA, 2WikiMultiHopQA, and MuSiQue are constructed among wikipedia or wikidata, via different multi-hop mining strategies with crowd-sourcing. They cover 3-5 hop questions with corresponding documents.
\end{itemize}

\paragraph{SFT Configurations} To construct the SFT dataset, we adopt \OurDataset{} (2k QA pairs) and employ DeepSeek-V3-0528 and DeepSeek-R1-0528 to generate teacher trajectories under the guidance of ground-truth reasoning chains, for the Instruct model and Thinking model respectively. We filter out samples with incorrect final answers and retain 1,536 instances. The input length in the SFT stage is set to 60K. Training is conducted for 4 epochs with a batch size of 256 and a learning rate of 1e-5.

\paragraph{Evaluation Configurations}
For all LLM backbones, we conduct evaluation under a 128K context window. Specifically, LLaMA3.1‑8B‑Instruct and Qwen2.5‑7B‑Instruct (with YaRN enabled) are evaluated with a maximum input length of 120K and an output limit of 10K. For all frontier models and the Qwen3 series, the input length is also set to 120K, with output limits of 10K for Instruct models and 30K for reasoning models.
For each question, we generate $N=4$ candidate responses and report the average score (Pass@1) in our main experiments, as well as Pass@k for test‑time scaling analyses. The Pass@k metric provides an unbiased estimate of the probability that at least one of the $k$ sampled responses is correct, given $n$ candidate solutions per problem. For multiple-choice tasks, we report standard accuracy. For open-end multi-hop QA tasks, we use the Hybrid Reward mentioned in Section~\ref{sec:advantage_reweight} and use GPT-OSS-120B as LLM-as-a-judge~\citep{LLM_as_a_Judge} to evaluate semantic equivalence between a model’s prediction and the ground-truth answer.

\paragraph{Training Prompt}
Here we list the training prompt template we used during Training.

\begin{tcolorbox}[colback=blue!6!white, colframe=gray!25!black, title=System Prompt during training]
You are a helpful assistant. Please read the provided text and answer the question below. Please structure your response into two main sections: Thought and Solution. In the Thought section, detail your reasoning process using the specified format: \verb|<begin_of_thought>| {thought with steps start with `Step N:'} \verb|<end_of_thought>| Each step should include detailed considerations such as analyzing questions, summarizing relevant findings, brainstorming new ideas, verifying the accuracy of the current steps, refining any errors, and revisiting previous steps. In the Solution section, based on various attempts, explorations, and reflections from the Thought section, systematically present the final solution that you deem correct. The solution should remain a logical, accurate, concise expression style and detail necessary step needed to reach the conclusion, formatted as follows: \verb|<begin_of_solution>| Therefore, the answer is \{insert answer here\} \verb|<end_of_solution>|.
    
\end{tcolorbox}

\paragraph{LLM-as-judge Prompt}
Here we list the prompts of LLM-as-Judge (1) when used as Outcome Reward Model to judge whether predicted answer is aligned with ground-truth answer; (2) when used as Sub-step Validity Signal to judge whether a sub-step in rollout trajectories aligns with the necessary entities and reasoning logic of the reference trajectory.

\begin{tcolorbox}[colback=blue!6!white, colframe=gray!25!black, title=Prompt for LLM-as-Judge as Reward Model]
You are an expert in verifying if two answers are the same.\\
Your input is a problem and two answers, Answer 1 and Answer 2. You need to check if they are equivalent.\\
Your task is to determine if two answers are equivalent, without attempting to solve the original problem.\\
Compare the answers to verify they represent identical values or meaning, even when written in different forms or notations.\\

Your output must follow the following format:\\
1) Provide an explanation for why the answers are equivalent or not.\\
2) Then provide your final answer in the form of: [[YES]] or [[NO]]\\

Problem: \textcolor[RGB]{158,160,161}{\{question\}}\\
Answer 1: \textcolor[RGB]{158,160,161}{\{predicted answer\}}\\
Answer 2: \textcolor[RGB]{158,160,161}{\{golden answer\}}
    
\end{tcolorbox}

\begin{tcolorbox}[colback=blue!6!white, colframe=gray!25!black, title=Prompt for LLM-as-Judge as Sub-step Validity Signal]
You are an expert in analyzing reasoning traces. Your task is to determine if a ``given reasoning substep from a model's output" is contained or reflected within the ``Ground Truth reasoning solution".\\

In your assessment, you must strictly adhere to the following special rules:\\
1. Ignore Step Order: You need to check if the logical content or core reasoning expressed by the substep is covered by the Ground Truth path, regardless of its position in either path.\\
2. Accept Varied Granularity: Differences in reasoning granularity are allowed. If the model substep is a logical combination (merger) of multiple steps in the Ground Truth, or if it is only a part (subset) of a single Ground Truth step, it should still be considered a match, as long as its core logic is clearly included or reflected in the Ground Truth path.\\

Specifically, you should check:\\
1. Does the substep text or its semantic equivalent appear in the Ground Truth solution?\\
2. Is the substep's core logic or reasoning step reflected or contained within the Ground Truth solution?\\
3. Does the substep represent a logical component that exists within the Ground Truth reasoning process?\\

You are checking if the substep EXISTS in the Ground Truth, not if it's correct or necessary for solving the problem.

Provide your final answer in the form of:

[[YES]] or [[NO]]\\

Ground Truth Reasoning Solution: \textcolor[RGB]{158,160,161}{\{ground truth\}}\\
Reasoning Substep to Check: \textcolor[RGB]{158,160,161}{\{substep\}}
    
\end{tcolorbox}

\section{Further Analysis}

\subsection{Generalization on Longer Context Window}

Although \OurMODEL{} is trained on a 60K input context window, we observe a strong generalization capacity to much longer contexts. As shown in Table~\ref{tab:length_generalization}, \OurMODEL{} achieve the strongest gains in the 16K–64K range—closest to the 60K training length. Notably, this improvement remains substantial even beyond 64K context window, while other baselines degrade sharply: \OurMODEL{} lifts LLaMA3.1‑8B‑Instruct from 34.62 to 48.08 and Qwen3‑4B‑Instruct from 51.92 to 57.69 on FRAMES with $>$64K input contexts. Upon questions with more than 128K contexts in Longbench V2, they achieve impressive absolute gains of +4.63\% and +6.48\%, respectively. Meanwhile, \OurMODEL{} exhibits strong performance on shorter input contexts ($<$16K). These results show that \OurMODEL{} not only attains optimal performance at its trained context length but also generalizes robustly to longer contexts, particularly on complex multi-hop reasoning tasks.

\begin{table*}[!t]
\centering
\caption{Overall \textbf{Pass@1} performance on long-context QA benchmarks. The top scores of each backbone LLM are \textbf{bolded}. Data in LongBench V2 is divided into three groups: Short ($<$32K), Medium (32K-128K), and Long ($>$128K).}
\label{tab:length_generalization}
\setlength\tabcolsep{6pt} 
\renewcommand{\arraystretch}{1.2} 
\fontsize{9pt}{10pt}\selectfont 

\resizebox{0.9\linewidth}{!}{
\begin{tabular}{lccccccc}
\toprule
\multicolumn{1}{c}{\multirow{2}{*}{\textbf{Models}}} & \multicolumn{4}{c}{\textbf{FRAMES}} & \multicolumn{3}{c}{\textbf{LongBench V2}} \\
\cmidrule(lr){2-5} \cmidrule(lr){6-8}

& 0-16K & 16K-32K & 32K-64K & $>$64K & Short & Medium & Long \\
\midrule
\specialrule{0em}{1pt}{1pt}\midrule
LLaMA3.1-8B-Instruct & 41.25 & 43.60 & 41.18 & 34.62 & 31.67 & \textbf{27.21} & 23.15 \\
- RLVR & 57.34 & 55.69 & 52.35 & 34.62 & 34.31 & 25.93 & 25.46 \\
\rowcolor{darkgreen!8}
- \textbf{\OurMODEL} & \textbf{60.62} & \textbf{60.57} & \textbf{62.65} & \textbf{48.08} & \textbf{34.86} & 26.16 & \textbf{27.78} \\
\midrule
Qwen3-4B-Instruct & 47.60 & 45.93 & 44.12 & 51.92 & 44.44 & 34.30 & 31.25 \\
- RLVR & 59.48 & 63.72 & 62.35 & 53.85 & 47.36 & \textbf{37.21} & 36.81 \\
\rowcolor{darkgreen!8}
- \textbf{\OurMODEL} & \textbf{63.39} & \textbf{67.68} & \textbf{66.47} & \textbf{57.69} & \textbf{53.47} & 36.74 & \textbf{37.73} \\
\midrule
Qwen3-4B-Thinking & 60.18 & 60.37 & \textbf{66.76} & 55.77 & 47.64 & 35.35 & 35.42 \\
- RLVR & 55.83 & 61.38 & 58.82 & 53.85 & 50.00 & 35.81 & \textbf{39.81} \\
\rowcolor{darkgreen!8}
- \textbf{\OurMODEL} & \textbf{63.12} & \textbf{67.38} & 66.47 & \textbf{63.46} & \textbf{52.08} & \textbf{36.05} & 38.43 \\
\bottomrule
\end{tabular}}
\end{table*}

\subsection{Analysis on the Quality of GT Reasoning Chains}
\begin{figure}[!t]
    \centering
    \includegraphics[width=0.7\textwidth]{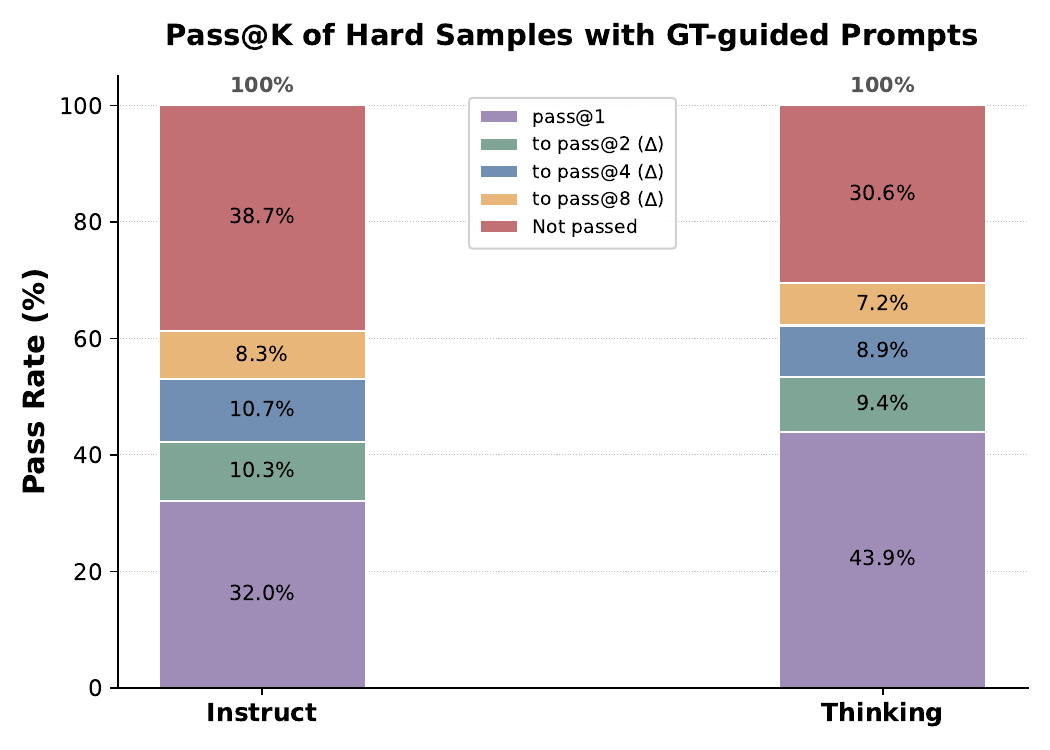}
    \vspace{-0.1cm}
    \caption{Pass@K performance of Qwen3‑4B models on hard samples using GT‑guided prompts. For each question in \OurDataset{}, we generate 8 rollouts and retain only those samples for which all predictions are incorrect (accuracy $= 0$).}
    \label{fig:hard_sample_pass_with_gt}
\end{figure}

During the Long‑Context QA Synthesis stage, we obtain multi‑hop QA pairs together with their corresponding multi‑hop reasoning chains. In this section, we validate the profound impact of providing explicit Ground Truth reasoning chains on solving high-difficulty multi-hop QA problems, which were previously unsolvable even after 8 rollouts (Pass@8 = 0). Figure~\ref{fig:hard_sample_pass_with_gt} clearly shows that the GT-guided prompts enabled LLMs to solve up to 69.4\% of these hard samples, confirming the effectiveness of this "step-level supervision" in long-text reasoning. Specifically, the combination of GT reasoning chains on Thinking model gains greater improvement, achieving a Pass@1 success rate of 43.9\% and reducing the total failure rate to 30.6\%. This demonstrates that the explicit reasoning chains effectively improve LLMs' answer accuracy on long-context multi-hop reasoning tasks, even on extremely hard questions, allowing the LLM to effectively trace the complex, multi-hop logic that connects the question to the reference answer.

\subsection{Observation of ``almost-there" phenomenon on training datasets}

\begin{figure*}[!t]
    \centering
    \subfloat[Entity Coverage]{\includegraphics[width=0.49\textwidth]{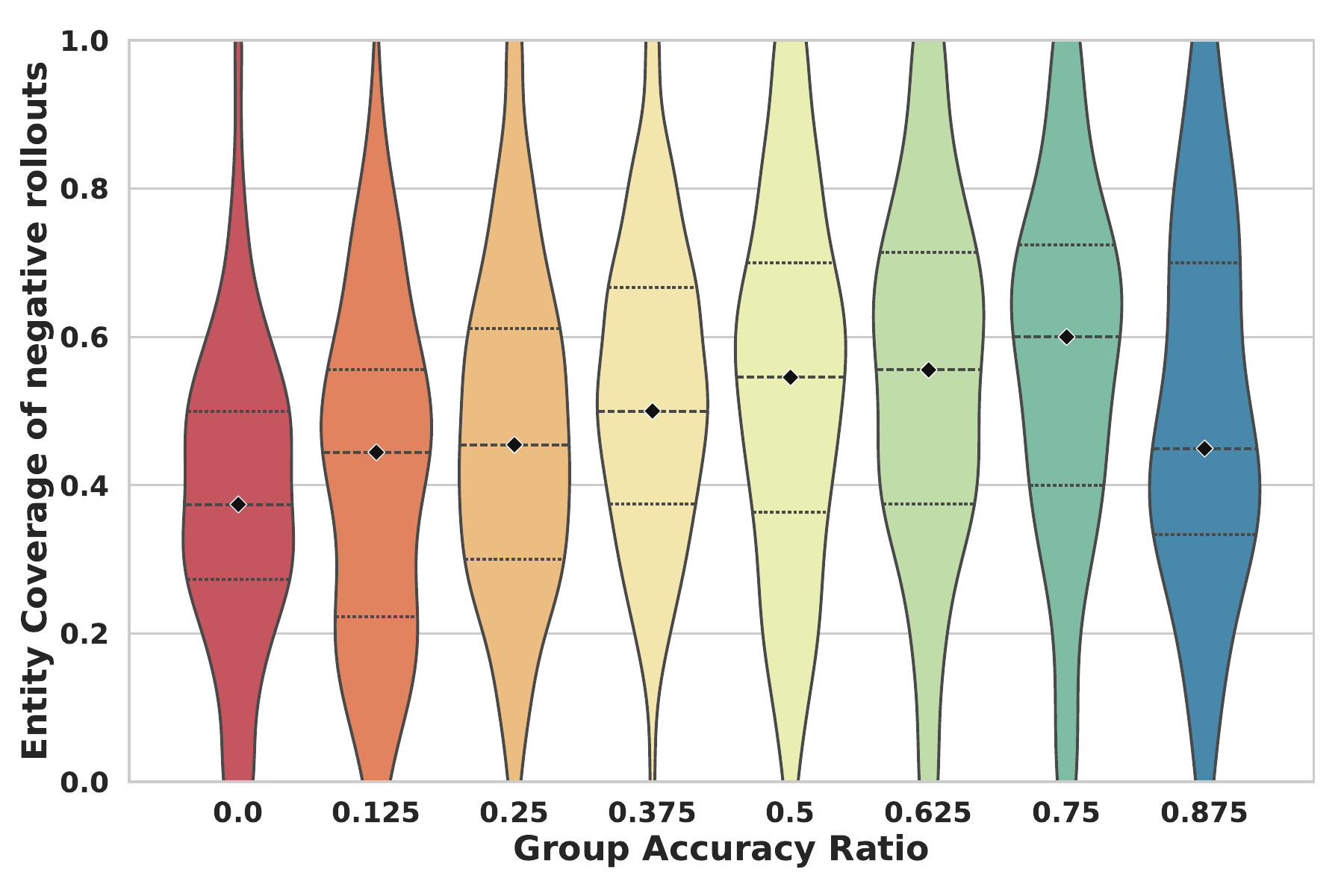}
    \label{entity_coverage_train_set}}  
    \hfill
    \subfloat[Triplet Coverage]{\includegraphics[width=0.49\textwidth]{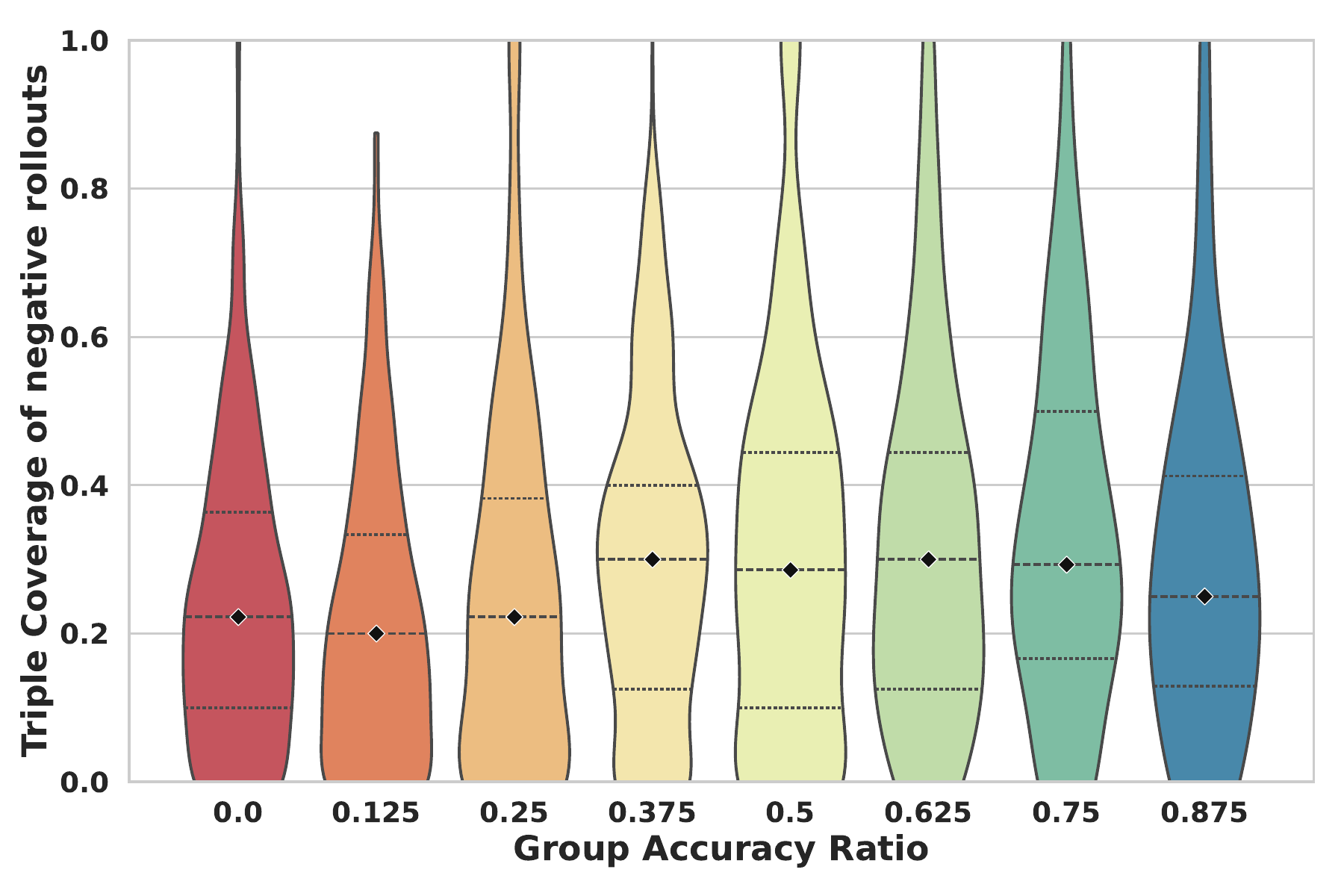}
    \label{triplet_coverage_train_set}}  
    \caption{Entity Coverage (a) and Triplet Coverage (b) distribution between negative rollouts and ground-truth reasoning chains on \OurDataset{} during training stage. Distributions are calculated according to average score in each group.}
    \label{fig:coverage_train_set}
\end{figure*}

In Section~\ref{sec:analysis_steps_in_negative}, we analyzed the commonly observed ``almost-there" phenomenon during RL training, where trajectories are largely correct but fail at the end due to a minor error. In this section, we further examine whether a similar Entity \& Triplet coverage trend appears on \OurDataset{}. As shown in Figure~\ref{fig:coverage_train_set}, both Entity Coverage and Triplet Coverage tend to increase as group accuracy improves. Moreover, the coverage scores peak and stay above the overall average when the Group Positive Ratio is around 50–75\%. This further demonstrates that LLMs achieve stronger information grounding but still fail at the subsequent combination and reasoning stage, underscoring the necessity of handling such ``almost-there" cases during RL training.

\subsection{Training Dynamics of Thinking Model}
\begin{figure}[!t]
    \centering
    \includegraphics[width=0.99\textwidth]{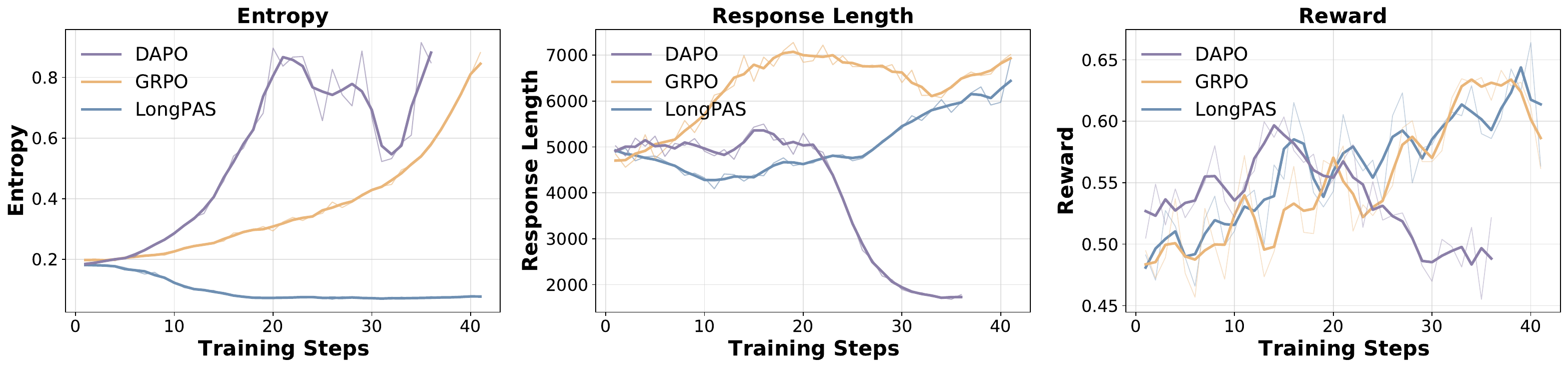}
    \vspace{-0.1cm}
    \caption{Training dynamics of \OurMODEL{} on Qwen3-4B-Thinking model compared with baseline algorithms. \textbf{Left}: Generation Entropy; \textbf{Middle}: Response Length; \textbf{Right}: Training Reward.}
    \label{fig:training_dynamics_thinking}
    \vspace{-0.25cm}
\end{figure}

We further analyze the step-by-step training dynamics of Qwen3-4B-Thinking model to further understand the training behaviors of \OurMODEL{} on Thinking model. As shown in Figure~\ref{fig:training_dynamics_thinking}, 

\subsection{Entity Coverage Dynamics}\label{app:entity_coverage}

\begin{wrapfigure}[16]{r}{0.5\textwidth}
    \centering
    \vspace{-0.4cm}
    {\includegraphics[width=0.48\textwidth]{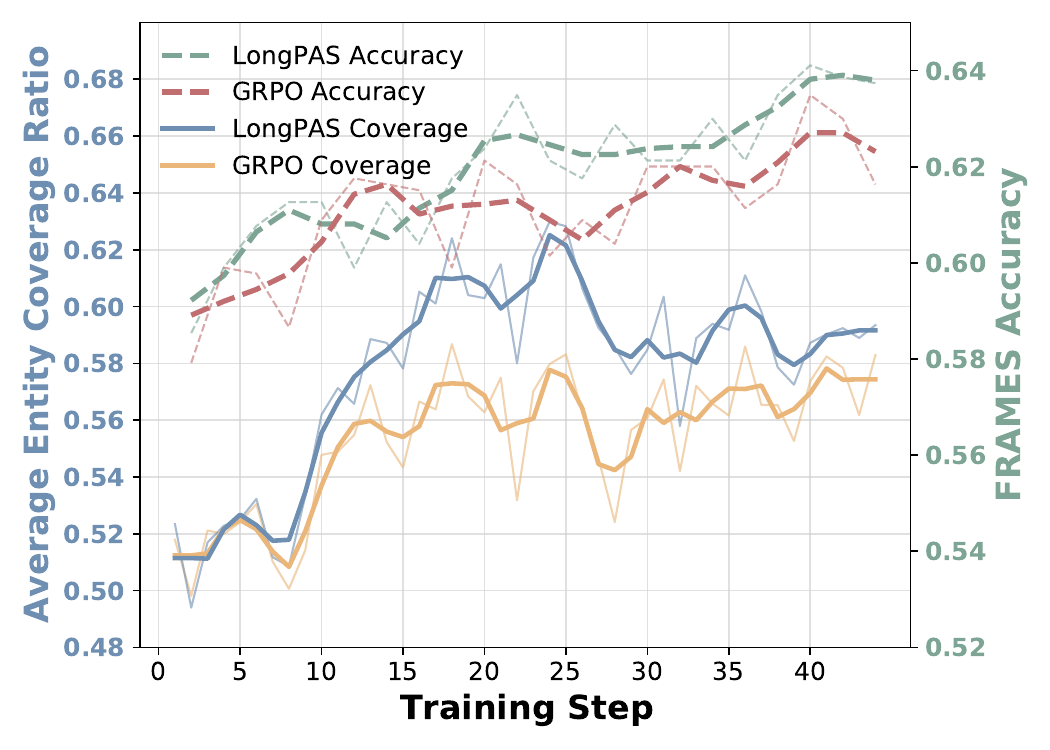}}
    \vspace{-0.2cm}
    \caption{Average Entity Coverage Ratio (\%) on the training data and FRAMES Accuracy dynamics with Qwen-4B model.}
    \label{fig:entity_coverage_dynamics}
\end{wrapfigure}

To further investigate the mechanism of \OurMODEL{} in mitigating the common wrong credit assignment confronting ``almost-there" samples, we analyze the dynamics comparison between average Entity Coverage Ratio and FRAMES Accuracy during training period. As shown in Figure~\ref{fig:entity_coverage_dynamics}, although both methods start at the same level, \OurMODEL{} quickly establishes and consistently maintains a clear advantage in the Average Entity Coverage Ratio. Its coverage curve stabilizes at a higher range around 0.60, notably above GRPO’s curve, which oscillates around 0.56. This indicates that \OurMODEL{} is more effective at grounding its reasoning in essential contextual information. Moreover, \OurMODEL{} attains a consistently higher peak in FRAMES accuracy compared with GRPO, demonstrating that its step‑level advantage shaping yields more precise credit assignment across the reasoning process, ultimately resulting in superior overall task performance.

\subsection{Test Time Scaling}
\begin{figure}[!h]
    \centering
    \includegraphics[width=0.99\textwidth]{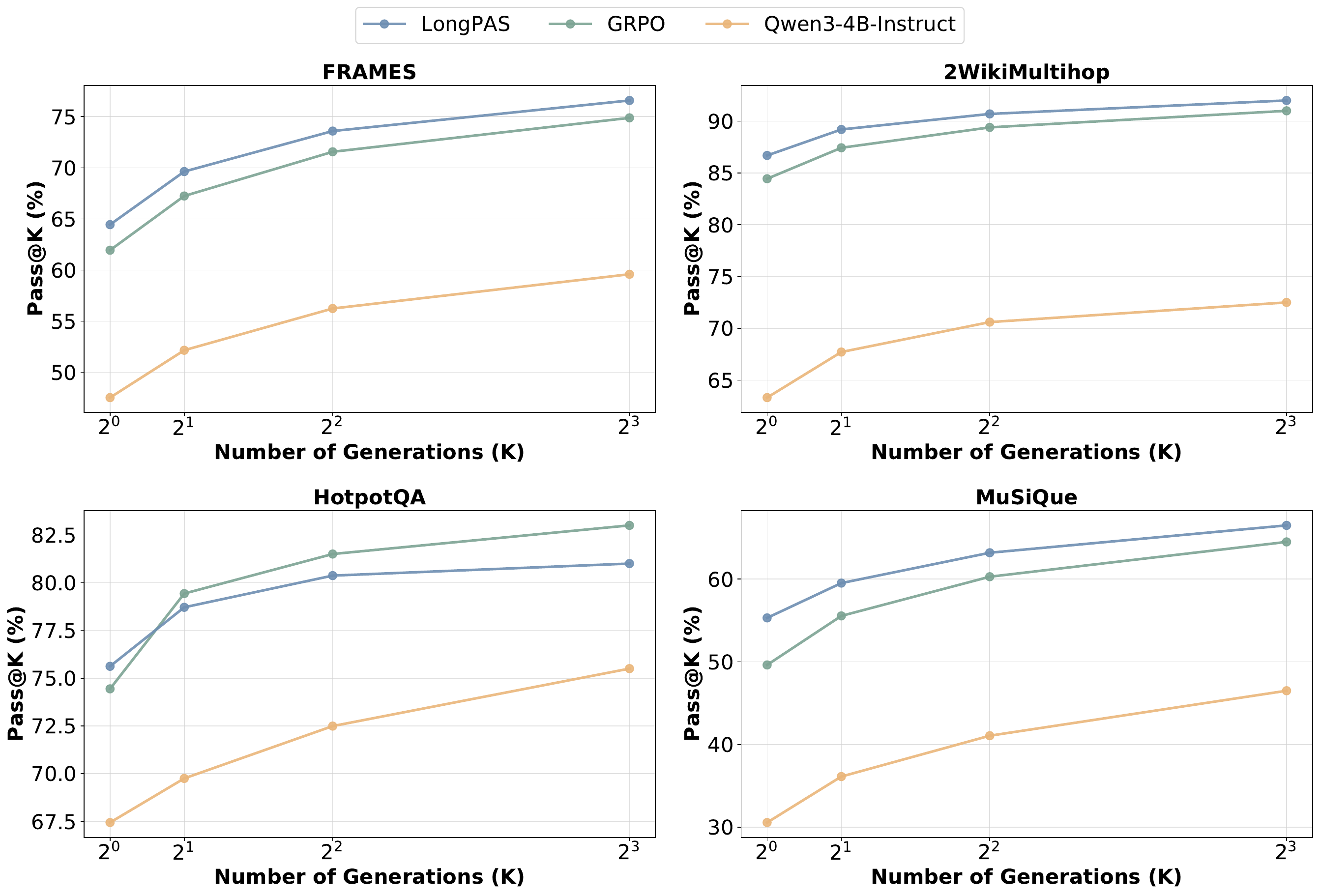}
    \vspace{-0.1cm}
    \caption{Test Time Scaling Performance (Pass@$k$) on four multi-hop reasoning benchmarks. The number of generations $k$ varies from 1 to 8.}
    \label{fig:tts}
    \vspace{-0.25cm}
\end{figure}

Prior studies have shown that with a limited number of rollouts, models often struggle to solve certain tasks, whereas a sufficiently large rollout budget substantially increases the probability of sampling effective solutions. Figure~\ref{fig:tts} reports the Pass@$k$ performance of \OurMODEL{}{} under test-time scaling settings. The results show that \OurMODEL{} achieves consistent gains as $k$ increases from 1 to 8. Notably, \OurMODEL{} also attains a higher Pass@1 score than Vanilla-GRPO, highlighting its effectiveness in boosting LLMs to produce precise reasoning processes for complex long-context multi-hop tasks during RL training.

\subsection{Detailed Results on LongBench V2}\label{sec:app_longbench_result}
We report the detailed results on each sub-task of LongBench V2 in Table~\ref{tab:app_longbench_result} to better illustrate the effectiveness of \OurMODEL{}.

\begin{table*}[t]
\centering
\caption{Overall performance of models on LongBench V2 benchmark. RLVR is implemented with GRPO~\citep{GRPO}. Results are reported using the \textbf{Avg@4} metric. The top scores for each backbone LLM are \textbf{bolded}.}
\label{tab:app_longbench_result}
\setlength\tabcolsep{6pt}
\renewcommand{\arraystretch}{1.2} 
\fontsize{9pt}{10pt}\selectfont 

\resizebox{\linewidth}{!}{
\begin{tabular}{l|cccccc|c}
\toprule
\multicolumn{1}{c|}{\textbf{Models}} & SingleDoc & MultiDoc & Code Repo & Dialogue & Long ICL & Long SDU & \textbf{Overall} \\
\midrule

\multicolumn{8}{c}{\cellcolor{gray!10}\textbf{\textit{Frontier Models}}} \\
\midrule
GPT5-Nano & 44.00 & 39.20 & 50.00 & 46.15 & 44.44 & 45.45 & 43.74 \\
Gemini-2.5-Flash-Thinking & 51.43 & 55.20 & 58.00 & 66.67 & 72.84 & 37.50 & 56.77 \\
GPT-OSS-120B & 44.57 & 43.20 & 53.06 & 61.54 & 46.91 & 48.48 & 47.01 \\
GPT-OSS-20B & 38.51 & 40.80 & 56.00 & 61.54 & 39.74 & 46.88 & 43.37 \\
\midrule

\multicolumn{8}{c}{\cellcolor{gray!10}\textbf{\textit{Instruct Models}}} \\
\midrule
LLaMA3.1-8B-Instruct & 28.86 & 26.20 & 27.00 & 32.05 & 26.23 & 30.30 & 27.93 \\
- RLVR & 31.86 & 29.20 & 29.00 & 32.69 & 23.46 & 21.21 & 28.93 \\
\rowcolor{darkgreen!8}
- \textbf{\OurMODEL} & 30.14 & 29.80 & 32.00 & 34.62 & 27.16 & 22.73 & \textbf{29.62} \\
\midrule
Qwen2.5-7B-Instruct & 36.71 & 28.20 & 32.00 & 35.90 & 36.42 & 30.30 & 33.60 \\
- RLVR & 34.43 & 29.00 & 30.00 & 40.38 & 25.31 & 25.00 & 31.01 \\
\rowcolor{darkgreen!8}
- \textbf{\OurMODEL} & 39.57 & 28.40 & 29.50 & 31.41 & 35.49 & 27.27 & \textbf{33.70} \\
\midrule
Qwen3-4B-Instruct & 36.00 & 30.60 & 36.50 & 60.26 & 41.67 & 32.58 & 37.28 \\
- RLVR & 39.14 & 37.00 & 49.50 & 60.90 & 43.21 & 40.91 & 42.10 \\
\rowcolor{darkgreen!8}
- \textbf{\OurMODEL} & 38.14 & 40.60 & 48.00 & 63.46 & 45.37 & 39.39 & \textbf{42.94} \\
\midrule
Qwen3-30B-A3B-Instruct & 42.43 & 38.40 & 52.00 & 62.82 & 47.53 & 37.21 & 44.43 \\
- RLVR & 50.29 & 50.40 & 42.00 & 56.41 & 43.21 & 36.36 & 47.91 \\
\rowcolor{darkgreen!8}
- \textbf{\OurMODEL} & 46.29 & 42.40 & 56.00 & 71.79 & 53.09 & 42.42 & \textbf{49.11} \\
\midrule

\multicolumn{8}{c}{\cellcolor{gray!10}\textbf{\textit{Reasoning Models}}} \\
\midrule
Qwen3-4B-Thinking & 37.00 & 35.60 & 41.50 & 60.26 & 44.44 & 42.42 & 40.46 \\
- RLVR & 37.14 & 40.80 & 46.00 & 66.67 & 41.98 & 33.33 & 41.75 \\
\rowcolor{darkgreen!8}
- \textbf{\OurMODEL} & 40.29 & 40.00 & 43.00 & 62.82 & 40.43 & 40.91 & \textbf{42.30} \\
\midrule
Qwen3-30B-A3B-Thinking & 44.00 & 44.00 & 46.00 & 64.10 & 55.56 & 54.55 & 48.31 \\
- RLVR & 40.00 & 46.40 & 50.00 & 51.28 & 49.38 & 33.33 & 44.53 \\
\rowcolor{darkgreen!8}
- \textbf{\OurMODEL} & 47.43 & 54.40 & 60.00 & 76.492 & 58.02 & 45.45 & \textbf{54.27} \\
\bottomrule
\end{tabular}}
\end{table*}

\subsection{Case Study}

To illustrate the qualitative differences in reasoning, we present a comparative case study using trajectories generated by Qwen3-4B-Thinking and \OurMODEL{} for the same question.

As shown in the examples below, the task involves multi-hop fact retrieval and date-based quantitative reasoning. The model must extract three pieces of information from different parts of the documents: (i) the person after whom the ship was named, (ii) the year the ship sank, and (iii) the birth and death years of that person, and then perform logical comparison and arithmetic calculation. The reasoning trajectory of \OurMODEL{} exhibits more advanced critical reasoning and temporal logic, covering stages such as Information Grounding, Information Extraction, Strategy Adjustment, Temporal Calculation, Self-Correction, and Answer Confirmation.

Comparing the two reasoning trajectories, we observe a shift from ``simple pattern matching and computation" to ``logical reasoning based on state judgment". The reasoning trajectory of \OurMODEL{} exhibits several richer reasoning patterns: (1) \textbf{Exhaustive Evidence Retrieval} – It performs more detailed and systematic information grounding, actively scanning and cross-checking relevant spans instead of relying on superficial matches; (2) \textbf{Constraint Consciousness} – It shows a stronger awareness of task boundaries and constraints throughout the thinking process; (3) \textbf{Robust Verification Loop} – It adopts a more divergent validation strategy, revisiting candidate answers, cross-validating them with multiple pieces of evidence, and rejecting inconsistent hypotheses before committing to a final conclusion.

We also observe that both models share an initial segment of correct reasoning steps, but the vanilla model later diverges into incorrect deductions and calculations. This phenomenon further demonstrates that \OurMODEL{} can better learn from ``almost-there" samples during RL training, leading to more accurate and precise reasoning.

\begin{tcolorbox}[breakable, colback=blue!5!white, colframe=blue!75!black, title=Case Trajectory on FRAMES with LongPAS-4B-Thinking]
\textbf{Question:} If the man that the SS Edmund Fitzgerald was named after was living at the time of the ship's sinking, how old was he? If he was already deceased, how long had he been dead? You may just use the year without regard for the date of birth.\\

\textbf{Ground Truth:} Edmund Fitzgerald, the man for whom the ill-fated ship SS Edmund Fitzgerald was named was born in 1895, he was 80 years old in 1975 when she sank.\\
\tcblower

\think{\\
I need to answer the question about how old Edmund Fitzgerald would have been if he was alive at the time of the ship's sinking, or how long he had been dead if he was already deceased. The question says I can just use the year without regard for the date of birth.\\

First, I need to find out who Edmund Fitzgerald was and when he was born and died. From the text provided, I can see that the ship was named after someone named Edmund Fitzgerald.\\

Let me look through the text to find information about Edmund Fitzgerald (the person, not the ship).\\

In the ``Name and launch" section, I find this information:\\

``Northwestern Mutual wanted to name the ship after its president and chairman of the board, Edmund Fitzgerald. Fitzgerald's own grandfather and all great uncles had themselves been lake captains, and his father owned the Milwaukee Drydock Company, which built and repaired ships. Fitzgerald had attempted to dissuade the naming of the ship after himself, proposing the names Centennial, Seaway, Milwaukee and Northwestern. The board was resolute, and Edmund abstained from voting; the 36 board members voted unanimously to name her the SS Edmund Fitzgerald."\\

This tells me that Edmund Fitzgerald was the president and chairman of the board of Northwestern Mutual Life Insurance Company. He was the person after whom the ship was named.\\

I need to find out his birth and death dates. The text doesn't explicitly state his birth and death dates. I might need to use external knowledge, but since this is a reading comprehension task, I should stick to the information provided in the text.\\

Let me search the entire text for any mention of Edmund Fitzgerald's age or death.\\

\textcolor[RGB]{158,160,161}{[Following thinking texts abbreviated]}
\\}

\thought{\\
Step 1: I need to identify who the SS Edmund Fitzgerald was named after. The question asks about ``the man that the SS Edmund Fitzgerald was named after." From the provided text, in the ``Name and launch" section, it states: ``Northwestern Mutual wanted to name the ship after its president and chairman of the board, Edmund Fitzgerald." \textbf{\underline{This confirms that the ship was named after Edmund Fitzgerald}}, who was the president and chairman of the board of Northwestern Mutual Life Insurance Company.\\

Step 2: The next step is to determine if this Edmund Fitzgerald was alive at the time of the ship's sinking (November 10, 1975) and, if so, his age or how long he had been dead. The question specifies that I can use just the year without regard for the date of birth, so I need to find his birth and death years from the text. I carefully scanned the entire provided text for any mentions of Edmund Fitzgerald's personal details, including age or lifespan.\\

Step 3: \textbf{\underline{In the ``See also" section of the text, I found relevant information}}: ``Edmund B. Fitzgerald (1926–2013), American businessman, and son of the namesake (1895–1986) of the ship." This clearly states that the namesake (the person the ship was named after) was born in 1895 and died in 1986. The phrase ``namesake (1895–1986)" directly indicates the birth and death years of the individual after whom the ship was named. This is the key information needed to answer the question, as it provides the exact years without ambiguity.\\

Step 4: Now I need to verify if Edmund Fitzgerald (the namesake) was alive during the ship's sinking in 1975. Since he died in 1986, \textbf{\underline{he was definitely alive in 1975}} (as 1975 is before 1986). The question asks for his age if he was living at the time of sinking. Using the year-based calculation (as instructed to ignore dates of birth), I subtract his birth year from the sinking year: 1975 - 1895 = 80. The text specifies to use the year without regard for the date of birth, so I don't need to consider whether his birthday had occurred in 1975—this simplifies the calculation to 80 years old.\\

Step 5: I double-checked for any potential errors or alternative interpretations. The text confirms Edmund Fitzgerald was alive in 1958 (when the ship was launched, as he was involved in the naming process), and since he died in 1986, he was alive during the 1975 sinking. No other parts of the text mention his age or death date, so the ``See also" section is the only reliable source here. The calculation 1975 - 1895 = 80 is straightforward and aligns with the question's instruction to use years only. There are no contradictions in the text, so this is accurate.\\
}

\truesolution{\\
Therefore, the answer is 80.\\
}

\end{tcolorbox}

\begin{tcolorbox}[breakable, colback=blue!5!white, colframe=blue!75!black, title=Case Trajectory on FRAMES with vanilla Qwen3-4B-Thinking-2507]
\textbf{Question:} If the man that the SS Edmund Fitzgerald was named after was living at the time of the ship's sinking, how old was he? If he was already deceased, how long had he been dead? You may just use the year without regard for the date of birth.\\

\textbf{Ground Truth:} Edmund Fitzgerald, the man for whom the ill-fated ship SS Edmund Fitzgerald was named was born in 1895, he was 80 years old in 1975 when she sank.\\
\tcblower

\think{\\
Let me carefully analyze the text to find information about Edmund Fitzgerald (the person after whom the ship was named) and his age or death date.\\

From the provided text, I can find this relevant information in the ``Name and launch" section:\\

``Northwestern Mutual wanted to name the ship after its president and chairman of the board, Edmund Fitzgerald. Fitzgerald's own grandfather and all great uncles had themselves been lake captains, and his father owned the Milwaukee Drydock Company, which built and repaired ships. Fitzgerald had attempted to dissuade the naming of the ship after himself, proposing the names Centennial, Seaway, Milwaukee and Northwestern. The board was resolute, and Edmund abstained from voting; the 36 board members voted unanimously to name her the SS Edmund Fitzgerald."\\

This tells me that Edmund Fitzgerald was the president and chairman of the board of Northwestern Mutual, and the ship was named after him.\\

I need to find out when Edmund Fitzgerald died (or if he was alive at the time of the sinking).\\

Let me look for information about Edmund Fitzgerald's death in the text.\\

I see a note at the end of the text that says: ``Edmund Fitzgerald may also refer to:"\\

And it lists several people with the same name:\\

- ``The Wreck of the Edmund Fitzgerald", a 1976 song about the shipwreck, by Gordon Lightfoot

- Edmund Fitzgerald Fredericks (1874/75–1935), Guyanese lawyer and Pan-African activist

- Edmund B. Fitzgerald (1926–2013), American businessman, and son of the namesake (1895–1986) of the ship

- The Edmund Fitzgerald (band), a UK math rock group\\

\textcolor[RGB]{158,160,161}{[Following thinking texts abbreviated]}
\\}

\thought{\\
Step 1: I need to identify who the SS Edmund Fitzgerald was named after and determine if he was alive at the time of the sinking (November 10, 1975) or had already died.\\

Step 2: From the text, I find that ``Northwestern Mutual wanted to name the ship after its president and chairman of the board, Edmund Fitzgerald." This confirms that the ship was named after a man named Edmund Fitzgerald who was the president and chairman of the board of Northwestern Mutual.\\

Step 3: I need to find information about when this Edmund Fitzgerald died. Looking at the end of the text, I see a section that says: ``Edmund B. Fitzgerald (1926–2013), American businessman, and son of the namesake (1895–1986) of the ship". This indicates that the ship's namesake (the man the ship was named after) was Edmund Fitzgerald who lived from 1895 to 1986.\\

Step 4: The ship sank on November 10, 1975. Since the ship's namesake died in 1986, \textcolor{red}{\underline{he had already been dead for 1986 - 1975 = 11 years by the time of the sinking.}} \textcolor[RGB]{158,160,161}{[Wrongly deduce the conclusion that one is ``Dead."]}\\

Step 5: I verify that the note specifically states ``the namesake (1895–1986) of the ship," which means the ship's namesake was Edmund Fitzgerald who died in 1986. \textcolor{red}{\underline{There is no indication in the text that this person was alive at the time of the sinking.}} \textcolor[RGB]{158,160,161}{[Unsuccessful self-correction.]}\\

Step 6: The question asks ``If the man that the SS Edmund Fitzgerald was named after was living at the time of the ship's sinking, how old was he? If he was already deceased, how long had he been dead?" Since the man was deceased (died in 1986), and the ship sank in 1975, \textcolor{red}{\underline{he had been dead for 11 years.}} \textcolor[RGB]{158,160,161}{[Miscalculation of years of death.]}\\

Step 7: I confirm that the question says ``You may just use the year without regard for the date of birth," so I don't need to calculate exact months or days. The calculation of 1986 - 1975 = 11 years is sufficient.\\
}

\falsesolution{\\
Therefore, the answer is \textcolor{red}{\textbf{\underline{11 years.}}}\\
}

\end{tcolorbox}

\section{Prompts \& Cases}

\subsection{QA Generation Prompts}
During question generation in Section~\ref{sec:QA_synthesis}, we employ specialized prompts to synthesize questions under different reasoning paradigms, including Multi-hop Reasoning, Temporal Reasoning, Causal Analysis and Hypothetical Scenarios. For better understanding, we showcase the detailed prompt used for multi-hop reasoning QA generation below:

\begin{tcolorbox}[colback=white, colframe=gray!50!black, title=Example of Multi-hop Reasoning QA Generation Prompt]
TASK: As a specialist in complex problem design, you will be provided with \{k\_context\} Wikipedia articles and a set of ``Multi-hop Reasoning Paths" derived from a knowledge graph. Your goal is to generate one challenging factoid ``Multihop Reasoning" question that meets the following criteria.\\

Your design principles are to maximize reasoning steps, maximize information lookup dependencies, and minimize direct clues while ensuring uniqueness and verifiability of the answer.\\

1. Core Requirements\\

* Standalone \& Context-Independent: Questions must not reference ``Article 1," ``Article 2," etc. They should be fully understandable without any additional context.\\
* Unambiguous \& Verifiable Answer: The question should have a single, clear, and factual answer. \\
* Chain-of-Thought Reasoning: The question must be structured to force a rigorous, step-by-step reasoning process. Each logical step must build upon the previous one, following the provided multi-hop reasoning path.\\

2. Advanced Reasoning Requirements\\

* Multi-hop Reasoning via Knowledge Graph Paths: Each question must be constructed by tracing and combining information along the provided multi-hop reasoning path. The path acts as a logical blueprint, connecting entities and concepts from different articles. It's important to note that the path may only involve a subset of the \{k\_context\} articles.\\
    * Multi-hop Reasoning Path Format: (Subject 1)-[Relation 1]-(Subject 2)-[Relation 2]-(Subject 3)-[Relation 3]-(Subject 4)...\\
* Generate a problem that requires reasoning through multiple entities and relationships. The problem should call for starting from one entity, reaching another through multiple chains of relationships, and analyzing the significance of this connection.\\

3. Output Format\\

For QA pair, follow this exact format:

[[Question]]:

[[Answer]]:

[[Explanation]]: Clearly explain the reasoning process. For each step, bullet point the specific piece of information (including the number/fact and the article it came from) used from the Wikipedia articles to formulate the question and its answer.\\

\textcolor[RGB]{158,160,161}{[N In-context documents and reasoning paths demonstrations abbreviated]}\\

\end{tcolorbox}

\subsection{QA Cases of \OurDataset{}}

In this section, we list the detailed cases of RL training data we constructed during Knowledge-Guided Long-Context Multi-hop QA Synthesis in Section~\ref{sec:QA_synthesis}, including Multi-hop Reasoning, Temporal Reasoning, Causal Analysis and Hypothetical Scenarios.

\begin{tcolorbox}[colback=green!5!white, colframe=green!75!black, title=Case of KG-guided Synthesis: \textbf{Multi-hop Reasoning}]
    \textbf{Question:}
    
    What European capital served as the city of exile for the head of a six-generation publishing house, a central figure in a Los Angeles Times award-winning debut novel, who shares the narrative with a Canadian academic? This academic's research focus on nostalgia and subsequent mental decline thematically links to the professional specialization of a comatose doctor from a canonical comic book. This comic's narrative begins in the same year that a future 'Savior' in a fantasy television series, then a homeless youth in a Midwestern U.S. state, was inspired by reading 'The Ugly Duckling' to choose her surname.
    
    \vspace{2mm}
    \textbf{Answer:}
    Vienna.

    \tcblower
    \textbf{Reasoning Chain:}\\
    (Emily Oliver)-[dismisses]-(Nádja)-[is a close friend of]-(John Price)-[is the protagonist of]-(Prague)-[deals with the history of]-(Horváth Kiadó)-[is the head of]-(Imre Horváth)-[was exiled in]-(Vienna)
\end{tcolorbox}

\begin{tcolorbox}[colback=green!5!white, colframe=green!75!black, title=Case of KG-guided Synthesis: \textbf{Temporal Reasoning}]
    \textbf{Question:}
    
         A Polish husband-and-wife sociological team introduced an English-language term for the meta-study of the scientific enterprise in a paper published in a year ending in 5. Exactly 11 years later, a quarterly journal dedicated to this field, but using a more versatile one-word term, was founded in their home country. Decades later, in a year divisible by 5, data was published indicating that a particular social science conferred a higher percentage of its doctorates on African-Americans than did a natural science, a field whose mathematical rigor is sometimes said to be a source of "envy" for the "softer" sciences. What is the duration, in years, between the founding of this one-word term quarterly journal and the publication of the Ph.D. distribution data?

    \vspace{2mm}
    \textbf{Answer:} 69 Year.

    \tcblower
    \textbf{Reasoning Chain:}\\
    (the Ossowscy)-[introduced the term]-(Science of science)-[is also called]-(Logology)-[is the study of]-(science)-[is mismatched with]-(economics)-[is an example of]-(softer sciences)-[overuse]-(mathematics)-[is used in]-(social sciences)-[is a type of]-(psychology)-[has a higher proportion of African-American Ph.D.s than]-(physics)
\end{tcolorbox}

\begin{tcolorbox}[colback=green!5!white, colframe=green!75!black, title=Case of KG-guided Synthesis: \textbf{Causal Analysis}]
    \textbf{Question:}
    
         In a novel first published in German in 1937, a fatal act of revenge by a Muslim protagonist against his Armenian rival is theorized to be a fictionalized account of the author's own youthful romantic frustrations. Correspondence from a town on the Amalfi Coast was instrumental in confirming the identity of this author, who wrote under a pseudonym. In an alternate timeline where the Central Powers triumphed in the global conflict of the 1910s, what was the resulting geopolitical stance of the nation where this author spent his final years?

    \vspace{2mm}
    \textbf{Answer:}
    Remained neutral through the entire war.
    
    \tcblower
    \textbf{Reasoning Chain:}\\
    (Nachararyan)-[is rival of]-(Ali)-[murders]-(Nachararyan)-[is rival for love of]-(Nino)-[was basis for]-(Zhenia Flatt)-[was teenage love interest of]-(Nussimbaum)-[was receiving income as]-(Kurban Said)-[is identified as]-(Essad Bey)-[wrote letter in]-(Positano)-[is located in]-(Italy)-[is neutral in]-(Great War)
\end{tcolorbox}

\begin{tcolorbox}[colback=green!5!white, colframe=green!75!black, title=Case of KG-guided Synthesis: \textbf{Hypothetical Scenario}]
    \textbf{Question:}
    
        Imagine a hypothetical scenario where a warlord from the Muromachi-Azuchi period, whose martial philosophy was famously summarized as "being crazy to die," is tasked with analyzing the ethical underpinnings of the celebrated story of the masterless warriors of the AkÅ Domain. While acknowledging their loyalty, this 16th-century figure would likely find their actions to be a departure from the more pragmatic, victory-focused ethos of his own era. Based on the historical development of the samurai moral code, what philosophical system, which became a required norm for samurai for the first time during the subsequent era of prolonged peace, would he identify as the primary influence that reshaped the warrior's way into a more refined moral and ethical theory?

    \vspace{2mm}
    \textbf{Answer:}
    Confucianism.

    \tcblower
    \textbf{Reasoning Chain:}\\
    (Chūshingura (A Treasury of Loyal Retainers))-[tells the story of]->(Forty-seven rōnin of the Akō Domain)-[were sentenced to]->(seppuku)-[is part of]->(Bushido)-[was influenced by]->(Confucianism)-[is related to]->(bushido)-[is related to]->(Hagakure)-[contains sayings attributed to]->(Nabeshima Naoshige)-[is representative figure of]->(Sengoku bushido)-[is from period]->(Muromachi-Azuchi (Sengoku period))
\end{tcolorbox}

\end{document}